\documentclass[graybox]{svmult}
\usepackage{amsmath,mathtools,amssymb}

\DeclarePairedDelimiter{\norm}{\lVert}{\rVert}
\usepackage{mathptmx}       
\usepackage{helvet}         
\usepackage{courier}        
\usepackage{type1cm}        
\usepackage{makeidx}         
\usepackage{graphicx}        
\usepackage{multicol}        
\usepackage[bottom]{footmisc}
\usepackage{acronym}
\usepackage{array}
\usepackage{booktabs}
\usepackage{multirow}
\makeindex             
\begin{document}

\title{Deep Learning and Reinforcement Learning for Autonomous Unmanned Aerial Systems: Roadmap for Theory to Deployment}
\titlerunning{Deep Learning and Reinforcement Learning for Autonomous UAS}
\author{Jithin Jagannath, Anu Jagannath, Sean Furman, Tyler Gwin}
\institute{Jithin Jagannath, Anu Jagannath, Sean Furman, Tyler Gwin \at Marconi-Rosenblatt Innovation Laboratory, ANDRO Computational Solutions, LLC, NY, USA \\\email{\{jjagannath, ajagannath, sfurman, tgwin\}@androcs.com}}
%
%
\maketitle

\abstract*{Each chapter should be preceded by an abstract (no more than 200 words) that summarizes the content. The abstract will appear \textit{online} at \url{www.SpringerLink.com} and be available with unrestricted access. This allows unregistered users to read the abstract as a teaser for the complete chapter.
Please use the 'starred' version of the \texttt{abstract} command for typesetting the text of the online abstracts (cf. source file of this chapter template \texttt{abstract}) and include them with the source files of your manuscript. Use the plain \texttt{abstract} command if the abstract is also to appear in the printed version of the book.}

\abstract{Unmanned Aerial Systems (UAS) are being increasingly deployed for commercial, civilian, and military applications. The current UAS state-of-the-art still depends on a remote human controller with robust wireless links to perform several of these applications. The lack of autonomy restricts the domains of application and tasks for which a UAS can be deployed. This is even more relevant in tactical and rescue scenarios where the UAS needs to operate in a harsh operating environment with unreliable wireless links. Enabling autonomy and intelligence to the UAS will help overcome this hurdle and expand its use improving safety and efficiency. The exponential increase in computing resources and the availability of large amount of data in this digital era has led to the resurgence of machine learning from its last winter. Therefore, in this chapter, we discuss how some of the advances in machine learning, specifically deep learning and reinforcement learning can be leveraged to develop next-generation autonomous UAS. \newline \indent
We first begin motivating this chapter by discussing the application, challenges, and opportunities of the current UAS in the introductory section. We then provide an overview of some of the key deep learning and reinforcement learning techniques discussed throughout this chapter. A key area of focus that will be essential to enable autonomy to UAS is computer vision. Accordingly, we discuss how deep learning approaches have been used to accomplish some of the basic tasks that contribute to providing UAS autonomy. Then we discuss how reinforcement learning is explored for using this information to provide autonomous control and navigation for UAS. Next, we provide the reader with directions to choose appropriate simulation suites and hardware platforms that will help to rapidly prototype novel machine learning based solutions for UAS. We additionally discuss the open problems and challenges pertaining to each aspect of developing autonomous UAS solutions to shine light on potential research areas. Finally, we provide a brief account of the UAS safety and regulations prior to concluding the chapter. 
}

\section{Introduction}
\label{sec:Intro}

The current era of \acp{UAS} has already made a significant contribution to civilian, commercial, and military applications \cite{gupta2013review, DoD}. The ability to have aerial systems perform tasks without having a human operator/pilot in the cockpit has enabled these systems to evolve in different sizes, forms, capabilities, conduct tasks, and missions that were previously hazardous or infeasible. Since the penetration of UAS into different realms of our lives is only going to increase, it is important to understand the current state-of-the-art, determine open challenges, and provide road-maps to overcome these challenges.  

The relevance of \acp{UAS} is increasing exponentially at a \ac{CAGR} of 15.5\% to culminate at USD 45.8 billion by 2025 \cite{UAV_Market}. While this growth seems extremely promising, there are several challenges that need to be overcome before UAS can achieve its full potential. The majority of these UASs are predominantly controlled by an operator and depend on reliable wireless communication links to maintain control and accomplish tasks. As the number of these systems increases and the mission complexity escalates, autonomy will play a crucial role in the next generation of UAS. In the next decade, we will see an incredible push towards autonomy for UAS just like how autonomy has evolved in markets like manufacturing, automotive industry, and in other robotics-related market areas.

When it comes to autonomy, there are several definitions and levels of autonomy claimed by manufacturers. Similarly, several definitions and requirements for various levels of autonomy exist in literature. According to \cite{Level_autonomy}, autonomy in UAS can be divided into five levels as follows,
\begin{itemize}
    \item \textbf{Level 1 - Pilot Assistance:} At this initial level, the UAS operator still maintains control of the overall operation and safety of the UAS. Meanwhile, the UAS can take over at least one function (to support navigation or maintaining flight stability) for a limited period of time. Therefore, at this level, the UAS is never in control of both speed and direction of flight simultaneously and all these controls are always with the operator. 
    \item \textbf{Level 2 - Partial Automation:} Here, the UAS is capable of taking control of altitude, heading, and speed in some limited scenarios. It is important to understand that the operator is still responsible for the safe operation of the UAS and hence needs to keep monitoring the environment and flight path to take control when needed. This type of automation is predominantly used for application with a pre-planned path and schedules. At this level, the UAS is said to be capable of \textit{sensing}.
    \item \textbf{Level 3 - Conditional Automation:} This case is similar to Level 2 described before with the exception that the UAS can notify the operator using onboard sensors if intervention is needed. This means the operator can be a little more disengaged as compared to Level 2 and acts as the backup controller. It is important to understand that at this level the scenarios of operation are relatively static. If any change in operating conditions is detected, the UAS will alert the operator to take over the control. At this level, the UAS is said to be capable of \textit{sense and avoid}.
    \item \textbf{Level 4 - High Automation:} At this level, the UAS is designed to operate without the requirement of the controller in several circumstances with the capability to detect and avoid obstacles using several built-in functionalities, rule sets, or machine learning-based algorithms deployed on the embedded computers on the UAS. While the operator can take control of the UAS, it is not necessary since several backup systems are in place to ensure safety in case one system fails. This is where an ideal system is expected to adapt to highly dynamic environments using powerful techniques like machine learning. At this level, the UAS is said to have achieved complete \textit{sense and navigate} capability.
    \item \textbf{Level 5 - Full Automation:} In this final level, the UAS operates fully autonomously without any intervention from operators regardless of the operating scenarios. This will not only include sense and navigate but the ability to learn and adapt its objectives and goals or even optimize its operational objectives and make necessary changes on-the-fly.  
\end{itemize}

Several of today's UASs have limited semi-autonomous modes (level 1 to 3) that warrants UAS to perform some autonomous actions such as return to the initial location, follow a pre-determined flight path, perform maneuvering acts, and recover from some standard instabilities, among others. A completely autonomous system (level 4 and 5) that can interact and survive in a dynamic environment without the need for human-in-the-loop are still far from being realized or deployed in a safe and effective manner. 

Machine learning, a subset of Artificial Intelligence has seen a spike in its application in various domains. This resurgence from its last winter is attributed to two main reasons (i) the exponential growth in computing resources in the last decade (ii) digitization of the modern era that has provided access to a huge quantity of data that can be used to train these machine learning models. Today, we see machine learning algorithms successfully applied to computer vision \cite{sebe2005machine,DL_CV,alexnet}, natural language processing \cite{NLP,NLP_2}, medical application \cite{litjens2017survey}, wireless communication \cite{JagannathAdHoc2019,Ajagannath6G2020}, signal intelligence \cite{Jagannath19MLBook}, robotics \cite{polydoros2017survey}, speech recognition \cite{DL_speech}, among others. These advancements in the field of machine learning have rendered it a perfect candidate to realize autonomy in UAS. To this end, in this chapter, we discuss the advances made in the field of machine learning, specifically deep learning, and reinforcement learning to facilitate autonomy to UAS. We also look at the key challenges and open research problems that need to be addressed for UAS autonomy. We hope this chapter becomes a great guide to beginners as well as seasoned researchers to take larger strides in these areas of research.

\subsection{Applications of UAS}

 \begin{figure}[h]
\centering
\includegraphics[width=4.7 in]{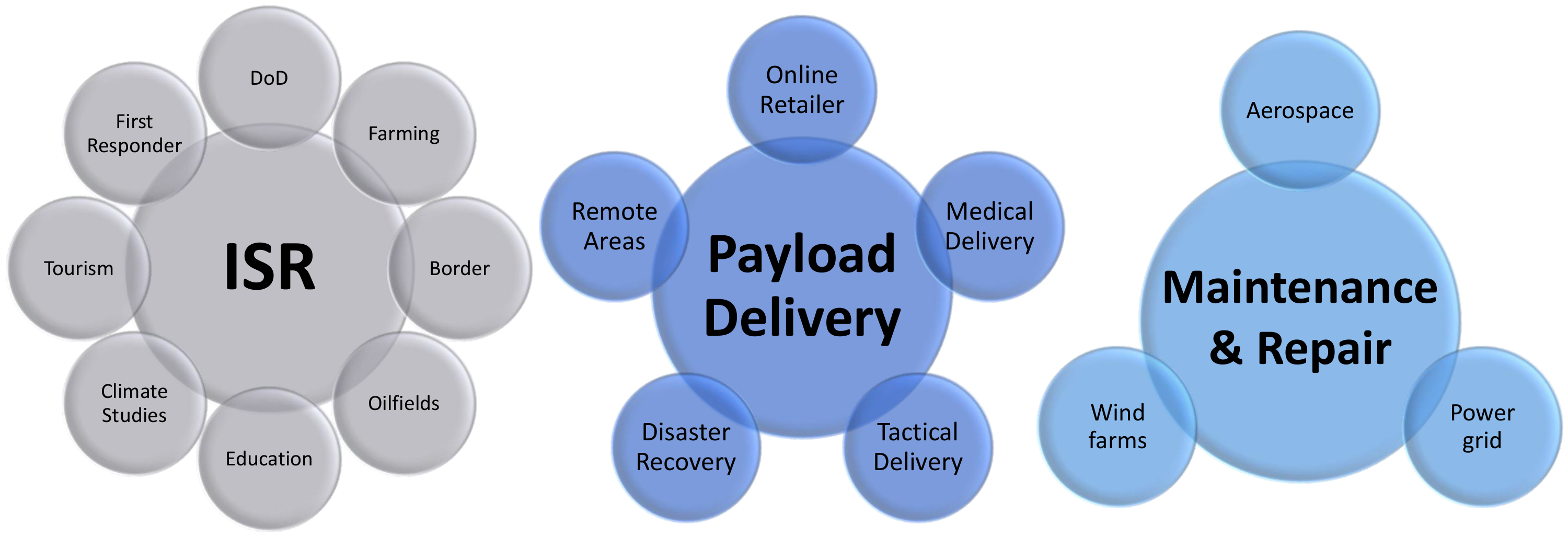}
\caption{Various applications of UAS}
\label{fig:App}       
\end{figure}

The applications of UAS can be broadly divided as follows, (i) \ac{ISR}, (ii) payload/product delivery, and (iii) maintenance and repair as shown in Figure \ref{fig:App}. Presently, ISR is the most common application that \acp{UAS} are employed for in both commercial and military realms. UASs are used for surveillance and remote sensing to map areas of interest using sensors such as traditional cameras or other sensors like acoustic, \ac{IR}, radars, among others. UASs are also used to monitor and survey oilfields, crop surveys, power grids, and other areas that are remote or difficult to access by operators. The surveys are also used for education, environment and climate studies, tourism, mapping, crop assessments, weather, traffic monitoring and border management. Similarly, UASs are used for humanitarian aid and rescue operations by first responders during disasters like flood and earthquakes where access by road does not exist or rendered inaccessible.

UAS is also being actively being designed and developed to become efficient agents for the delivery of payloads. These payloads include packages from online retailers, medical supplies to hospitals or areas of disaster, maintenance parts to remote locations in the commercial and civilian domains. As one can imagine delivery of different kinds of payloads will also be critical for several military missions and UAS might provide a safer alternative to accomplish such delivery in hostile areas with limited accessibility. Though not prevalent yet, it is envisioned that UAS will open up the market for several maintenance and repair tasks for the aerospace industry, power grid, wind farms, and other operations that are not easy to access. Currently, UAS are already being deployed to monitor and detect faults as well as provide maintenance alerts to reduce operational expense. It is envisioned that robotics enabled UAS will also be able to intervene when faults or necessary repair are detected in the near future.    

\subsection{Classification of UAS}
\label{subsec:classUAS}

\begin{figure}[h]
\centering
\hspace{-0.5 cm}
\includegraphics[width=4.8 in]{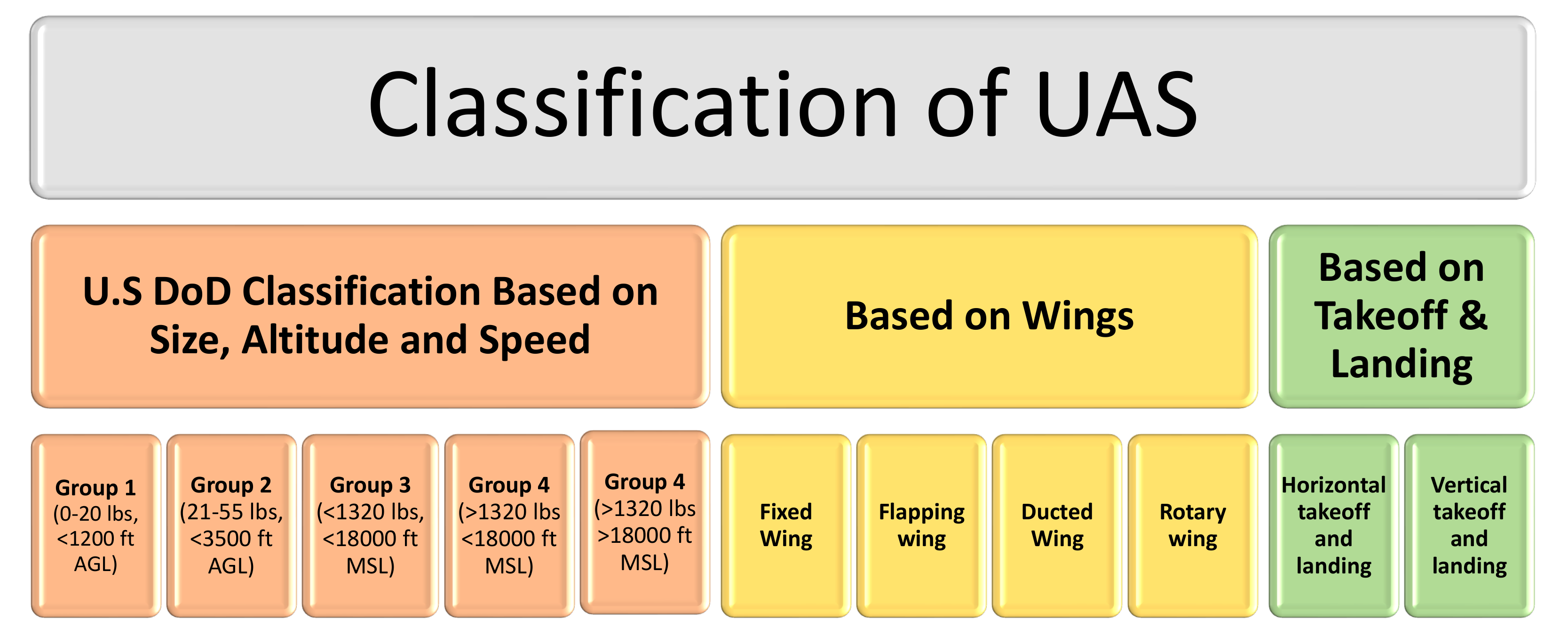}
\caption{Classification of Unmanned Aerial Systems}
\label{fig:Class}       
\end{figure}

It is clear from the previous discussion about the applications of UAS the need for versatility in the UAS design. Due to these reasons and the immense prospective that UAS holds for the future, UASs have evolved into different forms and sizes. While most of the discussion in this chapter is not specific to any type of UAS platforms, we provide a succinct classification for UASs in Figure \ref{fig:Class}. Several characteristics are used to classify different types of UASs. Here, we present the three most prevalent ones. The first is the classification of UASs adopted by \ac{DoD} which divides the systems into five groups based on weight, the altitude of flight, and the velocity. Another two sets of classification are based on the wing type and landing and takeoff. All these have been summarized in Figure \ref{fig:Class}.

\subsection{Chapter Organization}
\begin{figure}[h]
\centering
\includegraphics[width=4.6 in]{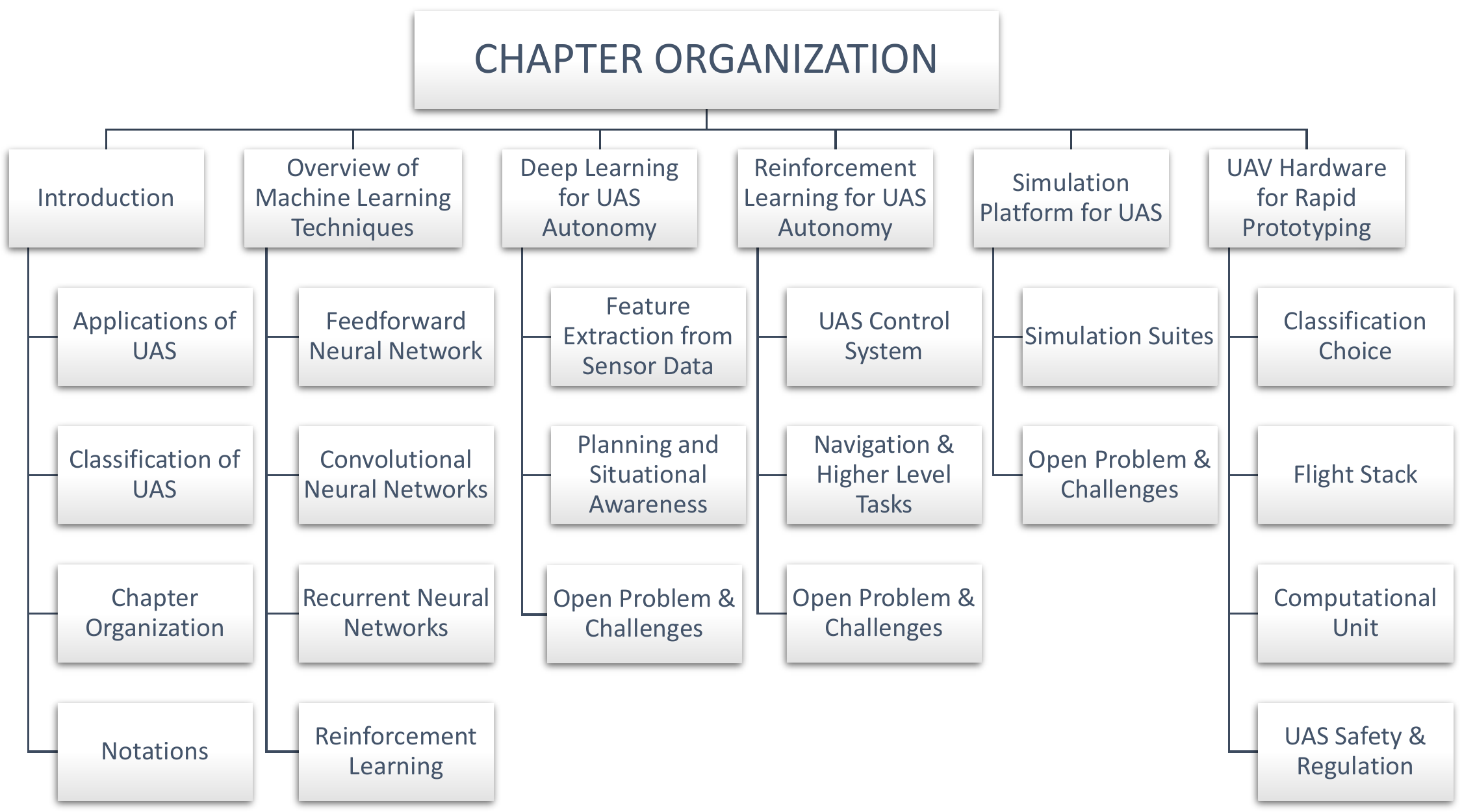}
\caption{Organization of the Chapter}
\label{fig:Organization}       
\end{figure}

This chapter is written for the benefit of a broad array of readers who have different levels of understanding and experience in this area of research. Therefore, for the benefit of readers who are relatively new to machine learning, we start by providing an overview of specific machine learning techniques that are explored in this chapter. The detailed explanation of these techniques would ensure even a beginner in the area of machine learning to grasp these techniques and benefit from the rest of the discussion in the chapter. The core contribution of this chapter is presented in the next four sections. Among these, two sections are dedicated to the discussion of various deep learning and reinforcement learning that has been explored for UAS. In each of these sections, we also discuss the open problems and challenges to motivate researches to explore these areas further. Since the goal of every research endeavor is to ensure the novel algorithms and solutions are effectively deployed on target platforms, in the next two sections, we look at simulation suites and hardware platforms that can help expedite this process. Finally, we conclude the chapter in the final section. The overall chapter organization is depicted in Figure \ref{fig:Organization}.

\subsection{Notations}

Here, we introduce some standard notations that will be used throughout this chapter. Matrices and vectors will be denoted by boldface upper and lower-case letters, respectively. For a vector $\mathbf{x}$, $x_i$  denotes the i-th element, $\norm{\mathbf{x}}$ indicates the Euclidean norm, $\mathbf{x}^\intercal$ represents its transpose, and $\mathbf{x} \cdot \mathbf{y}$ the Euclidean inner product of $\mathbf{x}$ and $\mathbf{y}$. For a matrix $\mathbf{H}$, $H_{ij}$ will indicate the element at row $i$ and column $j$. The notation $\mathbb{R}$ and $\mathbb{C}$ will indicate the set of real and complex numbers, respectively. The notation $\mathbb{E}_{x\sim p(x)}\left[f(x)\right]$ is used to denote the expected value, or average of the function $f(x)$ where the random variable $x$ is drawn from the distribution $p(x)$. When a probability distribution of a random variable, $x$, is conditioned on a set of parameters, $\boldsymbol{\theta}$, we write $p(x;\boldsymbol{\theta})$ to emphasize the fact that $\boldsymbol{\theta}$ parameterizes the distribution and reserve the typical conditional distribution notation, $p(x|y)$, for the distribution of the random variable $x$ conditioned on the random variable $y$. We use the standard notation for operations on sets where $\cup$ and $\cap$ are the infix operators denoting the union and intersection of two sets, respectively. We use $S_k \subseteq S$ to say that $S_k$ is either a strict subset of or equal to the set $S$ and $x \in S$ to denote that $x$ is an element of the set $S$. $\varnothing$ is used to denote the empty set and $|S|$ represents the cardinality of a set $S$. Lastly, the convolution operator is denoted as $*$. 

\section{Overview of Machine Learning Techniques}
\label{sec:Overview}

\begin{figure}[h]
\centering
\includegraphics[width=4.6 in]{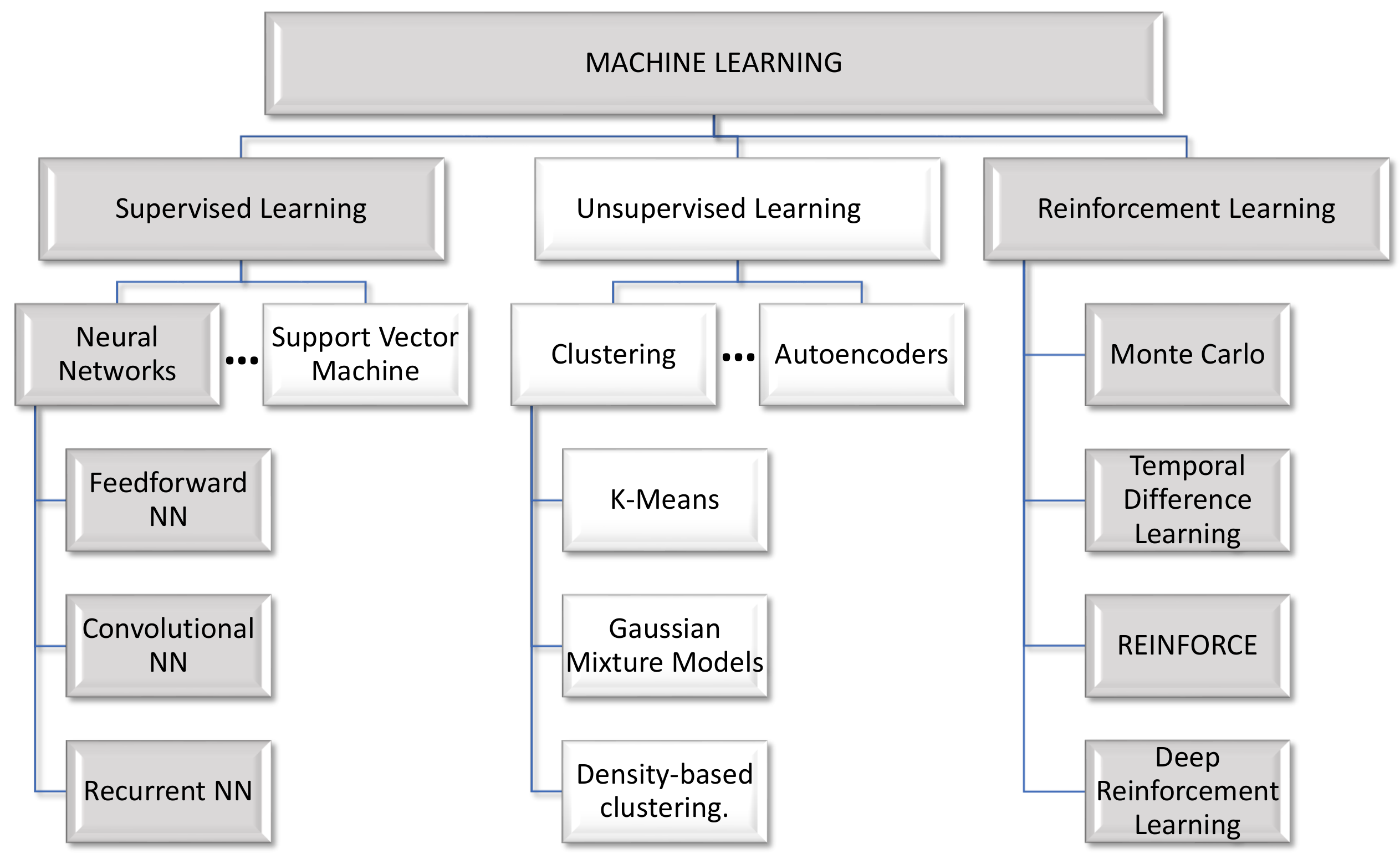}
\caption{Machine Learning Techniques}
\label{fig:1}       
\end{figure}

Machine Learning is a branch of artificial intelligence that is able to learn patterns from raw data and/or learn from observation sampling from the environment enabling computer systems to acquire knowledge. Machine learning is broadly classified into supervised, unsupervised, and reinforcement learning which are further subdivided into subcategories as shown in Fig.\ref{fig:1} (this is a very limited/relevant representation of this vast field). In this section, we elaborate on the key machine learning techniques (which are indicated as gray boxes in the  Fig.\ref{fig:1}) prominently used in this chapter to benefit readers in understanding the deep learning approaches for UAS autonomy.
\subsection{Feedforward Neural Networks}
\label{sec:fnn}
\begin{figure}[h]
\centering
\includegraphics[width=3.5 in]{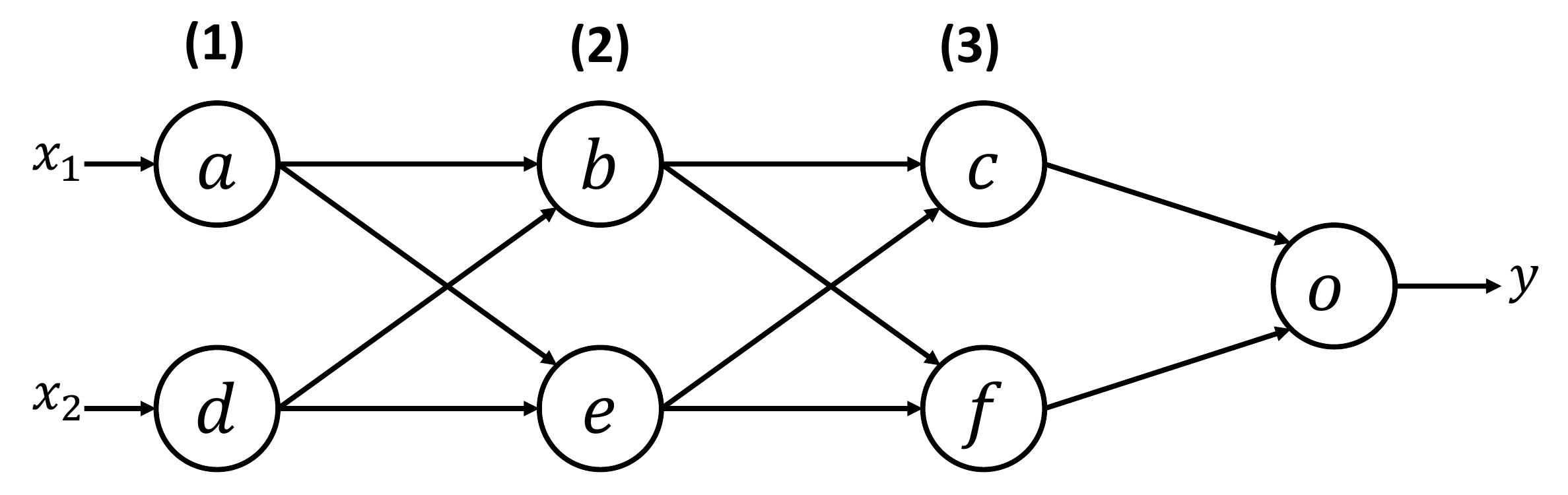}
\caption{Three-layered FNN}
\label{fig:fnn}  
\end{figure}

Feedforward neural networks (FNN) also referred to as multilayer perceptrons are directed layered neural networks with no internal feedback connections. Mathematically, an FNN performs a mapping, i.e., $f:X\longrightarrow Y$. An N-layered FNN is a composite function $y = f(\mathbf{x};\theta)=f_N(f_{N-1}(\cdots f_1(\mathbf{x})))$ mapping input vector $\mathbf{x}\in \mathbb{R}^m$ to a scalar output $y \in \mathbb{R}$. Here, $\theta$ represents the neural network parameters. The number of layers in the neural network dictates the \emph{depth} whereas the number of neurons in the layers defines the \emph{width} of the network. The layers in between the input and output layers for which the output does not show are called \emph{hidden} layers. Figure \ref{fig:fnn} shows a 3-layered FNN accepting a two-dimensional input vector $\mathbf{x}\in \mathbb{R}^2$ approximating it to a scalar output $y \in \mathbb{R}$.

In the figure, each node represents a neuron and each link between the nodes $i$ and $j$ are assigned a weight $w_{ij}$. The composite function of the 3-layered FNN is
\begin{equation}
    y = f(\mathbf{x};\theta) = f_3(f_2(f_1(\mathbf{x}))) \label{eq:fnn}
\end{equation}
In other words, the 3-layer FNN in Fig.\ref{fig:fnn} is the directed acyclic graph equivalent of the composite function in equation (\ref{eq:fnn}). The mapping in the first layer is 
\begin{equation}
    \mathbf{h}_1 = f_1(\mathbf{x}) = \mathcal{A}_1(\mathbf{W}_1\mathbf{x} + \mathbf{b}_1)
\end{equation}
where $\mathcal{A}_1(\circ)$ is the activation function, $\mathbf{b}_1$ is the bias vector, and $\mathbf{W}_1$ represents the weight matrix between the neurons in the first and second layers. Here, the weight matrix $\mathbf{W}_1$ is defined as the link weights between the neurons in the input and second layer
\begin{equation}
    \mathbf{W}_1 = \begin{bmatrix} w_{ab} & w_{db}\\w_{ae} & w_{de} \end{bmatrix}.
\end{equation}
Similarly, the second layer mapping can be represented as
\begin{equation}
    \mathbf{h}_2 = f_2(\mathbf{h}_1) = \mathcal{A}_2(\mathbf{W}_2\mathbf{h}_1 + \mathbf{b}_2)
\end{equation}
Finally, the output is
\begin{equation}
    y = f_3(\mathbf{h}_2) = \mathcal{A}_3(\mathbf{W}_3\mathbf{h}_2 + \mathbf{b}_3)
\end{equation}
The weight matrices in the second and final layers are
\begin{equation*}
    \mathbf{W}_2 = \begin{bmatrix} w_{bc} & w_{ec}\\w_{bf} & w_{ef} \end{bmatrix} \text{ and }
    \mathbf{W}_3 = \begin{bmatrix} w_{co} & w_{fo} \end{bmatrix}.
\end{equation*}
The neural network parameters $\theta = \{\mathbf{W}_1,\mathbf{W}_2,\mathbf{W}_3,\mathbf{b}_1,\mathbf{b}_2,\mathbf{b}_3 \}$ comprise the weight matrices and bias vectors across the layers. The objective of the training algorithm is to learn the optimal $\theta^*$ to get the target composite function $f^*$ from the available samples of $\mathbf{x}$.

\subsection{Convolutional Neural Networks}
\label{sec:cnn}
Convolutional networks or convolutional neural networks (CNNs) are a specialized type of feedforward neural network that performs convolution operation in at least one of its layers. The \emph{feature extraction} capability of CNNs mimics the neural activity of the animal visual cortex \cite{CNNcortex}. The visual cortex comprises a complex arrangement of cells that are sensitive to sub-regions of the perceived scene. The convolution operation in CNNs emulates this characteristic of the brain's visual cortex. Consequently, CNNs have been abundantly applied in the field of computer vision \cite{googlenet_inception,alexnet,LeNet5,vgg16,squeezenet,cnn4vision,fastRCNN,CNNface,resnet}. The convolution is an efficient method of feature extraction that reduces the data dimension and consequently reduces the parameters of the network. Hence, CNNs are more efficient and easier to train in contrast to its fully connected feedforward counterpart \ref{sec:fnn}. 

\begin{figure}[h]
\centering
\includegraphics[width=4.7 in]{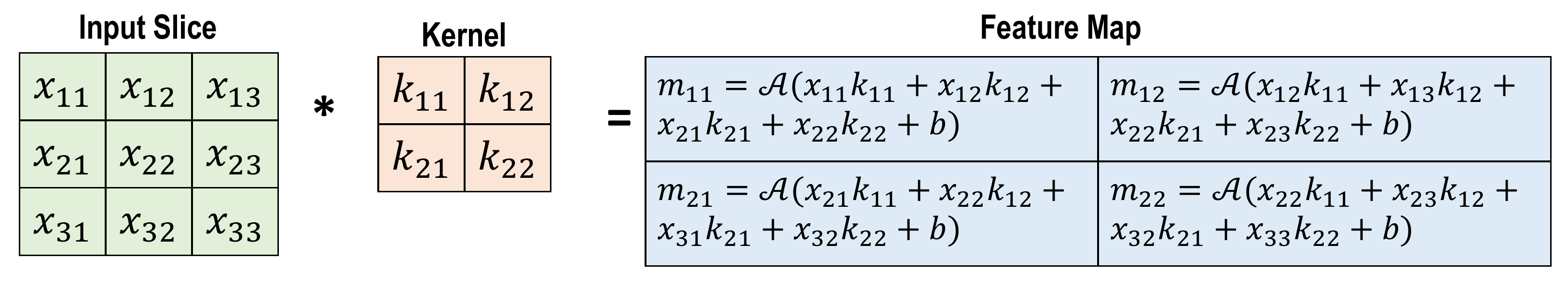}
\caption{Convolution of input slice with kernel}
\label{fig:cnn_conv}  
\end{figure}

A typical CNN architecture would often involve convolution, pooling, and output layers. CNNs operate on input tensor $\mathbf{X}\in \mathbb{R}^{W\times H \times D}$ of width $W$, height $H$, and depth $D$ which will be operated on by kernel (filter) $\mathbf{K}\in \mathbb{R}^{w\times h\times D}$ of width $w$, height $h$, and of the same depth as the input tensor to generate an output feature map $\mathbf{M}\in \mathbb{R}^{W_1\times H_1\times D_1}$. The dimension of the feature map is a function of the input as well as kernel dimensions, the number of kernels $N$, stride $S$, and the amount of zero padding $P$. Likewise, the feature map dimensions can be derived as $W_1 = \left(W-w+2P\right)/S + 1, \; H_1 = \left(H-h+2P\right)/S + 1,\; D_1 = N$.  Each kernel slice extracts a specific feature from the input region of operation. Kernel refers to the set of weights and biases. The kernel operates on the input slice in a sliding window manner based on the stride. Stride refers to the number of steps with which to slide the kernel along with the input slice. Hence, each depth slice of the input is treated with the same kernel or in other words, shares the same weights and biases - \emph{parameter sharing}. The convolution operation on an input slice $\mathbf{x}$ by a kernel $\mathbf{k}$ is demonstrated in Fig.\ref{fig:cnn_conv}. Here, $b$ represents the bias associated with the kernel slice and $\mathcal{A}\left(\circ\right)$ denotes a non-linear activation function.

\begin{figure}[h]
\centering
\hspace{-1 cm}
\includegraphics[width=2.8 in]{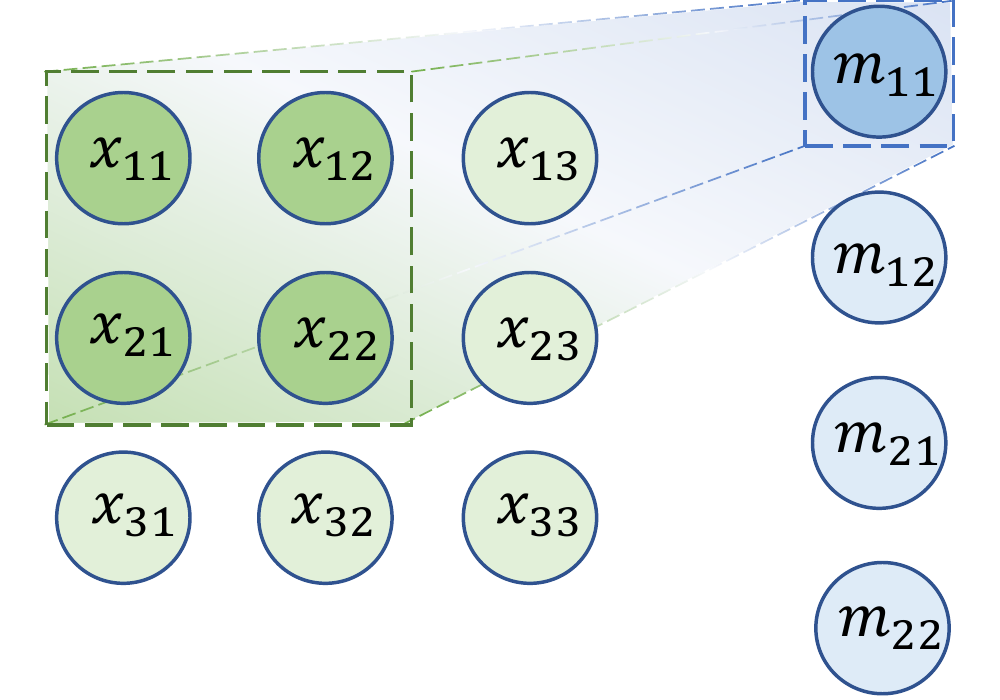}
\caption{Neural representation of convolution}
\label{fig:neural}  
\end{figure}

The resulting output from the convolution operation is referred to as the \emph{feature map}. Each element of the feature map can be visualized as the output of a neuron which focuses on a small region of the input - \emph{receptive field}. The neural depiction of the convolution interaction is shown in Fig.\ref{fig:neural}. 

\begin{figure}[h]
\centering
\includegraphics[width=2.8 in]{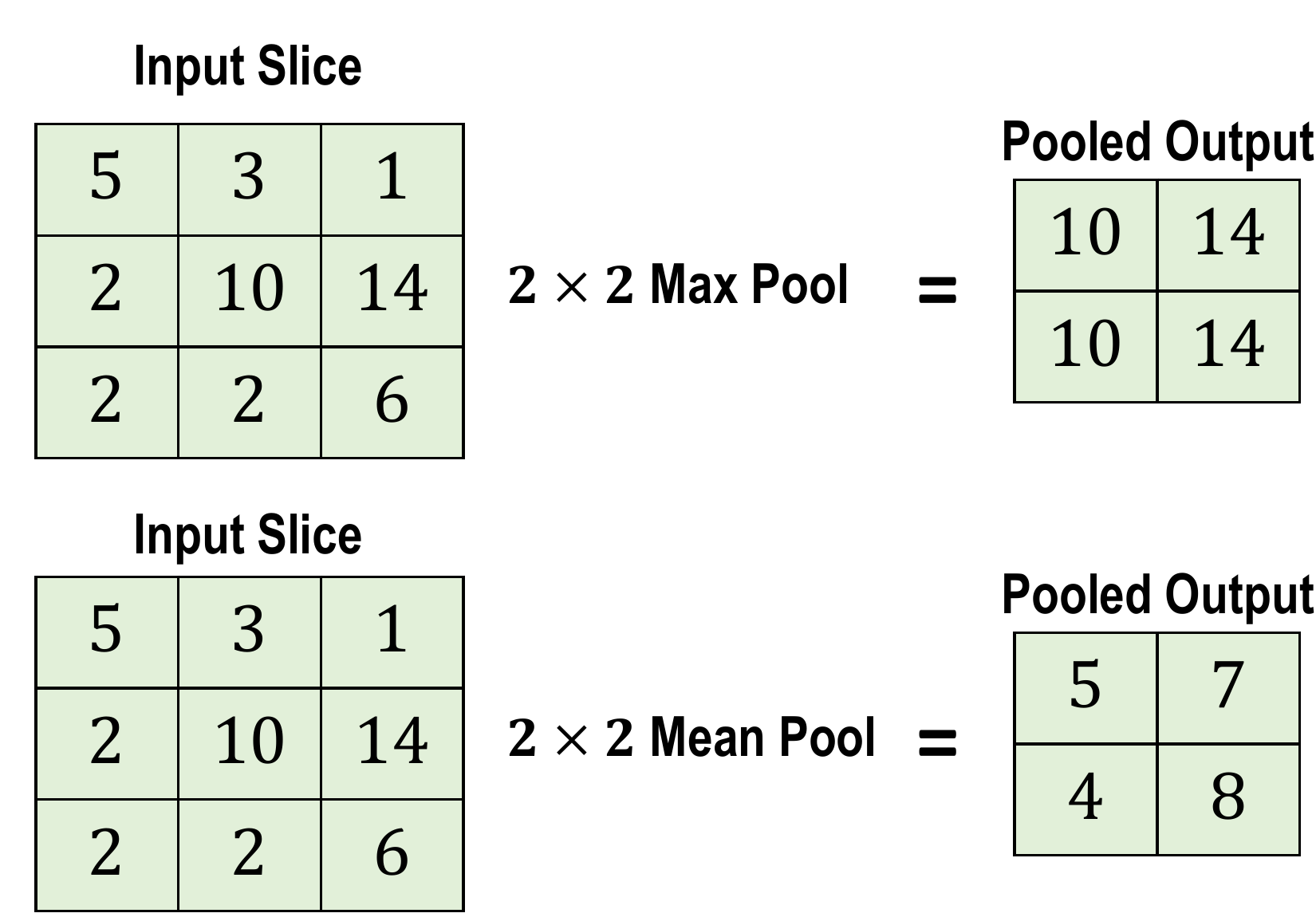}
\caption{Max and mean pooling on input slice with stride 1}
\label{fig:pool}  
\end{figure}

It is evident that each neuron in a layer is connected locally to the neurons in the adjacent layer - \emph{sparse connectivity}. Hence, each neuron is unaffected by variations outside of its receptive field while producing the strongest response for spatially local input pattern. The feature maps are propagated to subsequent layers until it reaches the output layer for a regression or classification task. \emph{Pooling} is a typical operation in CNN to significantly reduce the dimensionality. It operates on a subregion of the input to map it to a single summary statistic depending on the type of pooling operation - max, mean, $L_2$-norm, weighted average, etc. In this way, pooling downsamples its input. A typical pooling dimension is $2\times2$. Larger pooling dimensions might risk losing significant information. Figure \ref{fig:pool} shows max and mean pooling operations. 

A pooling layer of dimensions $W_p\times H_p$ upon operating over an input volume of size $W_1\times H_1\times D_1$ with a stride of $S_1$ will yield an output of volume $W_2 = \left( W_1-W_p\right)/S_1, \;H_2 = \left( H_1-H_p\right)/S_1, \; D_2 = D_1$. Pooling imparts invariance to translation, i.e., if the input to the pooling layer is shifted by a small amount, the pooled output will largely be unaffected \cite{Goodfellow-et-al-2016}.

As we have discussed, the three essential characteristics of CNNs that contribute to the statistical efficiency and trainability are parameter sharing, sparse connectivity, and dimensionality reduction. CNNs have demonstrated superior performance in computer vision tasks such as image classification, object detection, semantic scene classification, etc. Consequently, CNNs are increasingly used for UAS imagery and navigation applications \cite{uavapps}. Most notable CNN architectures are LeNet-5 \cite{LeNet5}, AlexNet \cite{alexnet}, VGG-16 \cite{vgg16}, ResNet \cite{resnet}, Inception \cite{googlenet_inception}, and SqueezeNet \cite{squeezenet}.

\subsection{Recurrent Neural Networks}
\label{sec:rnn}
\ac{RNN} \cite{Rumelhart1986} is a type of feedforward neural network specialized to capture temporal dependencies from sequential data. RNN holds internal memory states and recurrent connections between them to capture the sequence history. This characteristic of RNN enables it to exploit the temporal correlation of data rendering them suitable for image captioning, video processing, speech recognition, and natural language processing applications. Unlike CNN and traditional feedforward neural networks, RNN can handle variable-length input sequences with the same model. 

\begin{figure}[h]
\centering
\includegraphics[width=2 in]{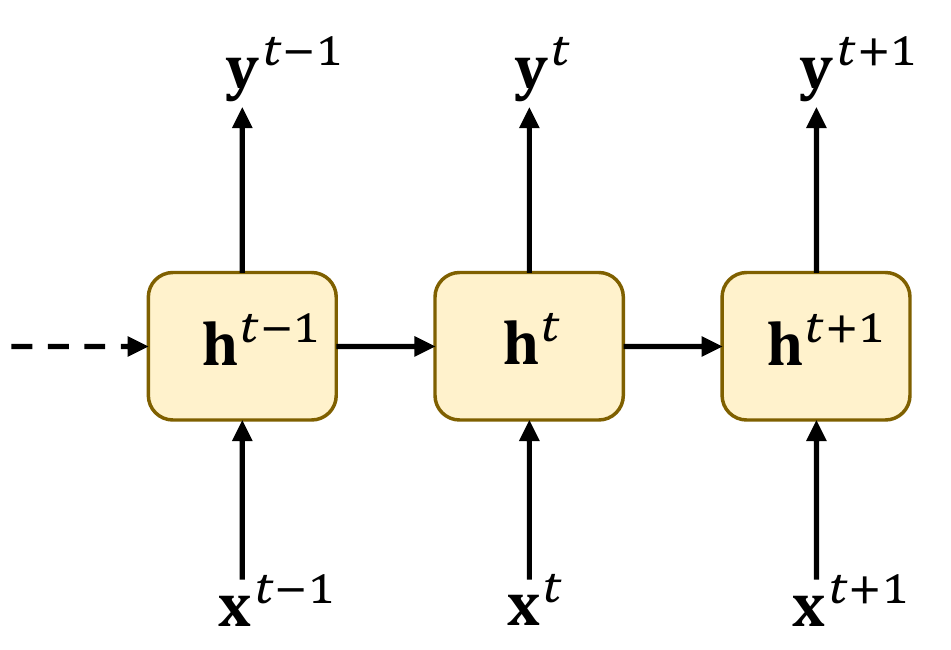}
\caption{Many-to-many RNN architecture}
\label{fig:mmrnn}  
\end{figure}

RNNs operate on input sequence vectors at varying time steps $\mathbf{x}^{t}$ and map it to output sequence vectors $\mathbf{y}^{t}$. The recurrence relation in an RNN parameterized by $\mathbf{\theta}$ can be expressed as 
\begin{equation}
    \mathbf{h}^t = \mathcal{F}\Big(\mathbf{h}^{t-1},\mathbf{x}^{t};\mathbf{\theta} \Big)
    \label{eq:recursive}
\end{equation}
where $\mathbf{h}^t$ represents the hidden state vector at time $t$. The recurrence relation represents a recursive dynamic system. By this comparison, RNN can be defined as \emph{a recursive dynamic system that is driven by an external signal, i.e, input sequence $\mathbf{x}^{t}$}. The equation (\ref{eq:recursive}) can be unfolded twice as
\begin{align}
  \mathbf{h}^t &= \mathcal{F}\Big(\mathbf{h}^{t-1},\mathbf{x}^{t};\mathbf{\theta} \Big)\\
  &= \mathcal{F}\Big(\mathcal{F}\Big(\mathbf{h}^{t-2},\mathbf{x}^{t-1};\mathbf{\theta} \Big),\mathbf{x}^{t};\mathbf{\theta} \Big)\\
  &= \mathcal{F}\Big(\mathcal{F}\Big(\mathcal{F}\Big(\mathbf{h}^{t-3},\mathbf{x}^{t-2};\mathbf{\theta} \Big),\mathbf{x}^{t-1};\mathbf{\theta} \Big),\mathbf{x}^{t};\mathbf{\theta} \Big)
\end{align}
The unfolded equations show how RNN processes the whole past sequences $\mathbf{x}^{t}, \mathbf{x}^{t-1},$ $\cdots, \mathbf{x}^{1}$ to produce the current hidden state $\mathbf{h}^{t}$. Another notable inference from the unfolded representation is the \emph{parameter sharing}. Unlike CNN, where the parameters of a spatial locality are shared, in an RNN, the parameters are shared across different positions in time. For this reason, RNN can operate on variable-length sequences allowing the model to learn and generalize well to inputs of varying forms. On the other hand, traditional feedforward network does not share parameters and have a specific parameter per input feature preventing it from generalizing to an input form not seen during training. At the same time, CNN share parameter across a small spatial location but would not generalize to variable-length inputs as well as an RNN. A simple many-to-many RNN architecture which maps multiple input sequences to multiple output sequences is shown in Fig.\ref{fig:mmrnn}.

\begin{figure}[h]
\centering
\includegraphics[width=4.8 in]{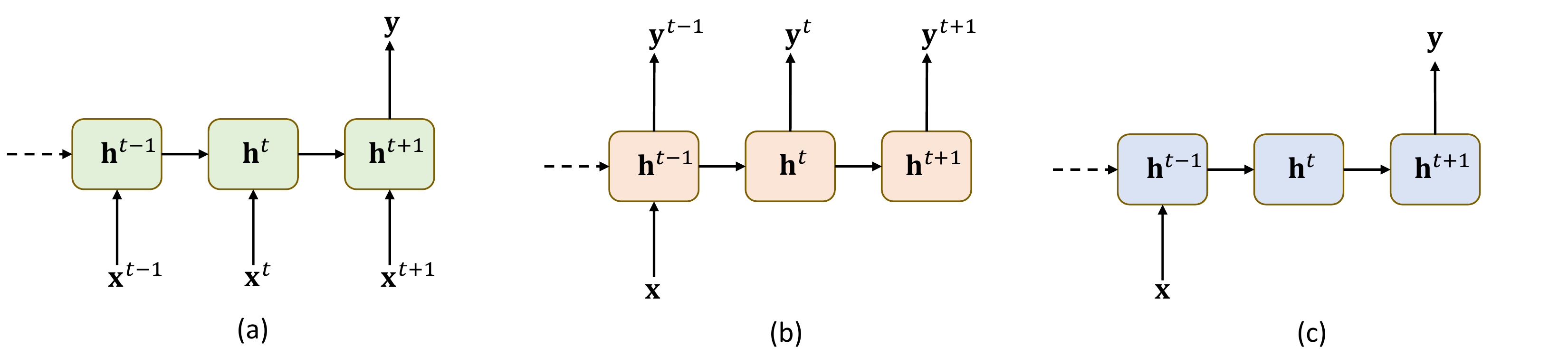}
\caption{RNN architectures. (a) Many-to-one, (b) One-to-many, and (c) One-to-one}
\label{fig:allrnn}  
\end{figure}

For a simple representation, let us assume the RNN is parameterized by $\mathbf{\theta}$ and $\mathbf{\phi}$ with input-to-hidden, hidden-to-hidden, and hidden-to-output weight matrices being $\mathbf{W}_{ih}, \mathbf{W}_{hh},$ and $\mathbf{W}_{ho}$ respectively. The hidden state at time $t$ can be expressed as 
\begin{align}
   \mathbf{h}^t &= \mathcal{F}\Big(\mathbf{h}^{t-1},\mathbf{x}^{t};\mathbf{\theta} \Big)\\ 
   &= \mathcal{A}_h\Big(\mathbf{W}_{hh}\mathbf{h}^{t-1} + \mathbf{W}_{ih}\mathbf{x}^{t} + \mathbf{b}_h\Big).
\end{align}
where $\mathcal{A}_h(\circ)$ is the activation function of the hidden unit and $\mathbf{b}_h$ is the bias vector. The output at time $t$ can be obtained as a function of the hidden state at time $t$,
\begin{align}
   \mathbf{y}^t &= \mathcal{G}\Big(\mathbf{h}^{t};\mathbf{\phi} \Big)\\
   &= \mathcal{A}_o\Big(\mathbf{W}_{ho}\mathbf{h}^t + \mathbf{b}_o\Big)
\end{align}
where $\mathcal{A}_o(\circ)$ is the activation function of the output unit and $\mathbf{b}_o$ is the bias vector. Other typical RNN architectures are shown in Fig.\ref{fig:allrnn}.

The RNN architectures discussed so far captures only hidden states from the past. Some applications would also require future states in addition to past. This is accomplished by a bidirectional RNN \cite{biRNN}. In simple words, bidirectional RNN combines an RNN that depends on past states (\emph{i.e.,} from $\mathbf{h}^{1}, \mathbf{h}^{2}, \mathbf{h}^{3}, \cdots, \mathbf{h}^{t}$) with that of an RNN which looks at future states (\emph{i.e.,} from $\mathbf{h}^{t}, \mathbf{h}^{t-1}, \mathbf{h}^{t-2}, \cdots, \mathbf{h}^{1}$).

\subsection{Reinforcement Learning}
\label{sec:ReinforcementLearning}
Reinforcement learning is focused on the idea of a goal-directed agent interacting with an environment based on its observations of the environment \cite{RL_book}. The main goal of reinforcement learning is for the agent to learn how to act i.e., what action to perform in a given environmental state, such that a reward signal is maximized. The agent repeatedly interacts with the environment in a series of discrete time steps by observing the environmental state, choosing, and executing an action. The action chosen by the agent may affect the state of the environment in the next time step. The agent receives a reward signal from the environment and transitions to a new state. The agent has some capability to sense the environmental state; informally the state can be thought of as any information about the environment that is made available to the agent. The agent selects which of the possible actions it can take by following a policy which is a function, in general stochastic, that maps state to actions. A reward signal is used to define the goal of the problem. The reward received by the agent at each time step specifies the immediate desirability of the current state. The objective of the reinforcement learning agent is to maximize the cumulative reward, typically defined by a value function, which defines the long-term goodness of the agent. The agent aims at achieving a goal by continuously interacting with the environment. This interaction which involves taking actions while trading off short and long term rewards renders reinforcement learning a potentially well-suited solution to many autonomous problems \cite{RL_Robotics_Survey_Kober}. 


The reinforcement learning problem is usually represented mathematically using a finite \ac{MDP}. A finite \ac{MDP} is defined by the following tuple $(S, A, P, R)$, where $S$, $A$, $P$, and $R$ are the state space, action space, transition function, and reward function respectively. Note that in finite \acp{MDP}, the state, action, and reward spaces consist of a finite number of elements. At each time step, the agent observes state $s \in S$, selects and takes action $a \in A$, receives a reward $r$, and transitions to a new state $s' \in S$. The transition function specifies the probability of transitioning from state $s$ to state $s'$ as a consequence of choosing action $a$ as,
\begin{equation}
    P(s,a,s')=Pr(S_{t+1}=s'|S_{t}=s,A_{t}=a).
\end{equation}
The reward function $R$ defines the expected reward received by the agent after transitioning to state $s'$ from state $s$ after taking action $a$ i.e., 
\begin{equation}
    R(s,a)=\mathbb{E}[R_t|S_t=s,A_t=a].
\end{equation}
It can be seen that the functions $P$ and $R$ define the dynamics of the \ac{MDP}. A reinforcement learning agent uses a policy to select actions in a given state. The policy, denoted $\pi(s,a)$ provides a probabilistic mapping of states to actions as,
\begin{equation}
    \pi(s,a)=Pr(A_t=a|S_t=s).
\end{equation}
As discussed earlier, value functions are used to define the long term goodness of the agent. Mathematically, the \emph{state-value function} is denoted as
\begin{equation}
    v_{\pi}(s)=\mathbb{E}_{\pi}\Bigg[\sum_{k=0}^{\infty}\gamma^kR_{t+k+1}|S_t=s\Bigg], \forall s\in S
\end{equation}
The \emph{state-value function} specifies the expected return, i.e., sum of discounted rewards, if the agent follows policy $\pi$ starting from state $s$. The discount rate $\gamma$, $0 \leq \gamma \leq 1$, is used to weight future rewards progressively less. For example, as $\gamma$ approaches zero the agent is concerned only with immediate rewards whereas when $\gamma$ approaches unity, the agent favors future rewards. The expected discounted return is denoted by $G_t$ i.e.,
\begin{equation}
    G_t=\sum_{k=0}^{\infty}\gamma^kR_{t+k+1}.
\end{equation}
Additionally, the \emph{action-value function} for policy $\pi$ is mathematically represented as 
\begin{equation}
    q_{\pi}(s,a)=\mathbb{E}_{\pi}\Bigg[\sum_{k=0}^{\infty}\gamma^kR_{t+k+1}|S_t=s,A_t=a\Bigg]
\end{equation}
The \emph{action-value function} specifies the expected return if the agent takes action $a$ in state $s$ under policy $\pi$. 

The \ac{MDP} dynamics of the environment and the notion of value functions have been exploited to develop multiple algorithms. In the case where the \ac{MDP} is fully known, i.e, the agent has knowledge of $P$ and $R$, dynamic programming methods (planning algorithms), such as policy iteration and value iteration can be used to solve the \ac{MDP} for the optimal policy or optimal value function. However, in reinforcement learning, knowledge of the \ac{MDP} dynamics is not usually assumed. Both model-based and model-free approaches exist for solving reinforcement learning problems. In model-based reinforcement learning, the agent attempts to learn a model of the environment directly, by learning $P$ and $R$, and then using the environmental model to plan actions using algorithms similar to policy iteration and value iteration. In model-free reinforcement learning, the agent does not attempt to directly learn a model of the environment but rather attempts to learn an optimal value function or policy. The discussion in this chapter is primarily focused on model-free methods. 

Generally speaking, model-free reinforcement learning algorithms fall into value function or policy gradient based methods. In value function based methods, the agent attempts to learn an optimal value function, usually action-value, and from which an optimal policy can be found. Value function methods include Monte Carlo, \ac{SARSA}, and Q-Learning. Policy gradient based methods attempt to learn an optimal parameterized policy directly via a gradient of a scalar performance measure with respect to the policy parameter. The REINFORCE algorithm is an example of a policy gradient method.

\subsubsection*{Monte Carlo}
Monte Carlo methods can be utilized to learn value functions and optimal policies by direct experience with the environment. In particular, sequences of states, actions, and rewards can be obtained by the agent interacting with the environment, either directly or in simulation, and the value function can be estimated by averaging the returns beginning from a state-action pair. Monte Carlo methods are typically used for episodic tasks. An episode (sequence of state, action, reward) is generated by the agent following policy $\pi$ in the environment and the value function estimate is updated at the conclusion of each episode. Monte Carlo methods can be used for control i.e., finding the optimal policy, by performing policy improvement. Policy improvement updates the policy such that it is greedy with respect to the current action-value function estimate. The greedy policy for an action-value function is defined such that for each state $s$ the action with the maximum action-value is taken i.e.,
\begin{equation}
    \pi(s)\doteq \operatorname*{argmax}_{a \in A} q(s,a).
\end{equation}
An important consideration for using Monte Carlo methods for value function prediction, and in reinforcement learning in general, is that of maintaining exploration. In order to learn the action-value function, all state-action pairs need to be explored. One way to achieve this is known as exploration whereby each episode begins in a particular state-action pair and all state-action pairs have a non-zero probability of being selected at the start of an episode. Exploration guarantees every state-action pairs will be visited an infinite number of times in the limit of an infinite number of episodes \cite{RL_book}. An alternative approach is to utilize a policy that allows for continued exploration. An example is the $\epsilon$-greedy policy in which most of the time an action (probability of $1-\epsilon$) is selected that maximizes the action-value function while occasionally a random action is chosen with probability $\epsilon$ i.e.,
\begin{equation}
\pi(s,a)=\begin{cases}
			1-\epsilon+\frac{\epsilon}{|A|}, & \text{if } a = a^*\\
            \frac{\epsilon}{|A|}, & \text{otherwise}
		 \end{cases}    
\end{equation}
There are two approaches to ensure continued exploration: on-policy and off-policy methods. In on-policy methods, the algorithm attempts to evaluate and improve the policy that is being used to select actions in the environment whereas off-policy methods are improving a policy different than the policy used to select actions. In off-policy methods, the agent attempts to learn an optimal policy, called the target policy, by generating actions using another policy that allows for exploration, called the behavior policy. Since the policy learning is from data collected ``off'' the target policy, the methods are called off-policy. Both on-policy and off-policy Monte Carlo control methods exist.

\subsubsection*{Temporal Difference Learning}
\ac{TD} learning defines another family of value function based reinforcement learning methods. Similar, to Monte Carlo methods, \ac{TD} learns a value function via interaction with the environment. The main difference between \ac{TD} and Monte Carlo is that \ac{TD} updates its estimate of the value function at each time step rather than at the end of the episode. In other words, the value function update is based on the value function estimate of the subsequent state. 
The idea of updating value function based on the estimated return $(R_{t+1}+\gamma V(S_{t+1}))$ rather than the actual (complete) reward as in Monte Carlo is known as bootstrapping.
A simple \ac{TD} update equation for value function is
\begin{equation}
    V(S_t)=V(S_t)+\alpha[R_{t+1}+\gamma V(S_{t+1})-V(S_t)]
\end{equation}
where $\alpha$ is a step size parameter. In the above equation, it is seen that the \ac{TD} method updates the value function estimate at the next time step. The target value for the \ac{TD} update becomes $R_{t+1}+\gamma V(S_{t+1})$ which is compared to the current value function estimate $(V(S_t))$. The difference between the target and the current estimate is known as the \ac{TD} error i.e.,
\begin{equation}
    \delta_{t}\doteq R_{t+1}+\gamma V(S_{t+1})-V(S_t)
\end{equation}
It can be seen that an advantage of \ac{TD} methods is its ability to update value function predictions at each time step which enables online learning. 

\textbf{SARSA:} is an example of an on-policy \ac{TD} control algorithm. The \ac{TD} update equation presented above is extended for action-value function prediction yielding the \ac{SARSA} action-value update rule as,
\begin{equation}
    Q(S_t,A_t) \leftarrow Q(S_t,A_t)+\alpha[R_{t+1}+\gamma Q(S_{t+1},A_{t+1})-Q(S_t,A_t)].\label{eq:sarsa}
\end{equation}
As shown in the equation (\ref{eq:sarsa}), the update is performed after each sequence of $(\cdots,S_t,A_t,R_{t+1},S_{t+1},A_{t+1,\cdots})$ which leads to the name \ac{SARSA}. It is to be noted that the $Q$ estimate is updated based on the sample data generated from the behavior policy $(R_{t+1}+\gamma Q(S_{t+1},A_{t+1}))$. For the control algorithm perspective, a greedy policy like $\epsilon$-greedy is often used. 

\textbf{Q-Learning:} is an off-policy \ac{TD} control algorithm and its update rule is given below.
\begin{equation}
    Q(S_t,A_t) \leftarrow Q(S_t,A_t)+\alpha[R_{t+1}+\gamma\max_{a}Q(S_{t+1},a)-Q(S_t,A_t)] \label{eq:qlearn}
\end{equation}
As an off-policy method, the learned action-value function estimate $Q$ is attempting to approximate the optimal action-value function $Q^*$ directly. This can be seen in the update equation (\ref{eq:qlearn}) where the target value is $R_{t+1}+\gamma\max_{a}Q(S_{t+1},a)$ compared to $R_{t+1}+\gamma Q(S_{t+1},A_{t+1})$ of \ac{SARSA}. Unlike in SARSA, the $Q$ value is updated based on the greedy policy for action selection rather than the behavior policy. SARSA does not learn the optimal policy but rather learns the action-values resulting from the $\epsilon$-greedy action selections. However, Q-learning learns the optimal policy resulting from the $\epsilon$-greedy action selections causing the online performance to drop occasionally \cite{RL_book}. 

The \ac{TD} methods can be further generalized with $n$-step bootstrapping methods which are an intermediate between Monte Carlo and \ac{TD} approaches. The $n$-step methods generalize the \ac{TD} methods discussed earlier by utilizing the next $n$ rewards, states, and actions in the value or action-value function updates. 


The value function based approaches discussed so far have been presented as tabular methods. The algorithms are tabular because the state-value or action-value function is represented as a table or an array. In many practical problems of interest, the state spaces are very large and it becomes intractable to learn optimal policies using tabular methods due to the time, data, and memory requirements to populate the tables. Additionally with massive state spaces, it is typical that the agent will enter states that are previously unseen requiring the agent to generalize from experiences in similar states. An example of an overwhelmingly large state space occurs when the environmental state is represented as a camera image; for example, an 8-bit, 200x200 pixel RGB image results in $256^{3*200*200}$ possible states. To cope with these challenges, optimal policies can be approximated by utilizing function approximation techniques to represent value functions and policies. The different function approximation techniques used in supervised learning can be applied to reinforcement learning. The specific use of deep neural networks as a means for function approximation is known as \ac{DRL} and is discussed later in this section. 

When using function approximation techniques, parameterized state-value or action-value functions are used to approximate value functions. A state-value estimate can be denoted as $\hat{v}(s;\boldsymbol{w}) \approx v_{\pi}(s)$ and an action-value estimate as $\hat{q}(s,a;\boldsymbol{w}) \approx q_{\pi}(s,a)$ where $\boldsymbol{w} \in \mathbb{R}^{d}$ is the parameter vector.  In principle, any supervised learning method could be used for function approximation. For example, a value function estimate could be computed using techniques ranging from a linear function of the state and weights to nonlinear methods such as an \ac{ANN}. \ac{SGD} and its variants are often used to learn the value of the parameter vectors.


\subsubsection*{REINFORCE}
In contrast to value function based approaches, policy gradient methods attempt to learn an optimal parameterized policy directly without the requirement of learning the action-value function explicitly. The policy that is learned is defined as
\begin{equation}
    \pi(a|s,\boldsymbol{\theta})=Pr(A_t=a|S_t=s,\boldsymbol{\theta}_t=\boldsymbol{\theta})
\end{equation}
which specifies the probability that action $a$ is taken at step $t$ in state $s$ and is parameterized by the vector $\boldsymbol{\theta} \in \mathbb{R}^m$. Policy gradient methods learn the value of the policy parameter based on the gradient of a performance measure $J(\boldsymbol{\theta})$ with respect to the parameter. In the episodic case, the performance measure can be defined in terms of the state value function assuming the episode starts from an initial state $s_0$ as
\begin{equation}
    J(\boldsymbol{\theta})\doteq v_{\pi_{\boldsymbol{\theta}}}(s_0)
\end{equation}
REINFORCE is an example of a policy gradient algorithm and is derived from the policy gradient theorem
\begin{equation}
    \nabla J(\boldsymbol{\theta}) \propto \sum_{s}\mu(s)\sum_{a}q_{\pi}(s,a)\mathbf{\nabla_{\theta}}\pi(a|s,\boldsymbol{\theta})
\end{equation}
where $\mu(s)$ is a distribution over states and the gradients are column vectors with respect to parameter vector $\boldsymbol{\theta}$. The policy gradient theorem provides an expression for the gradient of the performance measure with respect to the parameter vector. From the policy gradient theorem, the following equation is derived for the gradient of $J(\boldsymbol{\theta})$
\begin{equation}
    \nabla J(\boldsymbol \theta) = \mathbb{E}_{\pi}\Bigg[G_t\frac{\mathbf{\nabla_{\theta}} \pi(A_t|S_t,\mathbf \theta)}{\pi(A_t|S_t, \theta)}\Bigg]
\end{equation}
Using \ac{SGD}, the REINFORCE update rule for the policy parameter vector $\mathbf \theta$ can be derived as
\begin{equation}
    \mathbf{\theta}_{t+1} = \mathbf{\theta}_{t} + \alpha G_t \frac{\mathbf{\nabla_{\theta}} \pi(A_t|S_t,\mathbf{\theta}_t)}{\pi(A_t|S_t,\mathbf{\theta}_t)}.\label{eq:rein_up} 
\end{equation}
The update equation (\ref{eq:rein_up}) moves the parameter in a direction that increases the probability of taking action $A_t$ during future visits to the state $S_t$ in proportion to the return $G_t$. This causes the parameter to favor actions that produce the highest return. The normalization prevents choosing actions with a higher probability that may not actually produce the highest return. 



It is possible to generalize the policy gradient theorem and REINFORCE update rule with the addition of a baseline for comparison to the action values or returns. The baseline can be an arbitrary function or random variable. The motivation behind the use of a baseline is to reduce the variance in policy parameter updates.  The update rule for the REINFORCE algorithm with a baseline is given as
\begin{equation}
    \mathbf{\theta}_{t+1} = \mathbf{\theta}_{t} + \alpha (G_t-b(S_t)) \frac{\mathbf{\nabla_{\theta}} \pi(A_t|S_t,\mathbf{\theta}_t)}{\pi(A_t|S_t,\mathbf{\theta}_t)} 
\end{equation}
where $b(S_t)$ is the baseline. A common baseline is an estimate of the state-value $\hat{v}(S_t,\mathbf{w})$ parameterized by the weight vector $\mathbf{w} \in \mathbb{R}^l$. The idea of using a state-value function as a baseline can be extended with actor-critic methods. In actor-critic methods, a state-value function, called a critic, is utilized to assess the performance of a policy, called an actor. The critic introduces a bias to the actor's gradient estimates which can substantially reduce variance.

The two most recent policy gradient methods are \ac{TRPO} and \ac{PPO}. \ac{TRPO} was introduced in \cite{schulman2015trust} in order to prevent drastic policy changes by introducing an optimization constraint - Kullback-Leibler (KL) divergence. The policy is updated based on a trust-region and the KL constraint ensures that the policy update is not too far away from the original policy. The inclusion of KL constraint in the optimization problem introduces computational and implementation difficulty. However, \ac{PPO} introduced in \cite{schulman2017proximal} mitigates this implementation hurdle by incorporating the constraint term within the objective function. PPO computes the probability ratio between new and old policies. There are two variants of PPO - PPO with KL penalty and PPO with clipped objective. In the first variant, the KL constraint is introduced as a penalty term in the objective function such that it computes a policy update that does not deviate much from the previous policy while minimizing the cost function. In the second variant, the KL divergence is replaced with a clipped objective function such that the advantage function will be clipped if the probability ratio lies outside a range, say $1\pm\phi$. In contrast to TRPO, PPO is simpler to implement and tune.
\subsubsection*{Deep Reinforcement Learning}

Deep Reinforcement Learning is a popular area of current research that combines techniques from deep learning and reinforcement learning \cite{Arulkumaran_2017}. In particular, deep neural networks are used as function approximators to represent action-value functions and policies used in traditional reinforcement learning algorithms. This is of particular interest for problems that involve large state and action spaces that become intractable to represent using tabular methods or traditional supervised learning function approximators. A key capability of deep learning architectures is the ability to automatically learn representations (features) from raw data. For example, a deep neural network trained for image classification will automatically learn to recognize features such as edges, corners, etc. The use of deep learning enables policies to be learned in an end-to-end fashion, for example, learning control policies directly from raw sensor values. A famous exemplary deep reinforcement learning algorithm is the deep Q-Network that pairs Q-Learning with a deep \ac{CNN} to represent the action-value function \cite{mnih2013playing}. The deep Q-Network was able to achieve super human performance on several Atari games by using only visual information, reward signal, and available actions i.e., no game specific information was given to the agent. The deep Q-Network employs two methods to address the known convergence issues \cite{Tsitsiklis_analysis_td} that can arise when using neural networks to approximate the $Q$ function. These methods are experience replay and the use of a separate target network for $Q$ updates. The experience replay mechanism stores sequences of past experiences, $(s_t,a_t,s_{t+1},r_{t+1})$, over many episodes in replay memory. The past experiences are used in subsequent $Q$ function updates which improve data efficiency, removes correlations between samples, and reduces the variance of updates. The separate target network $\hat Q$ is used for generating targets in the Q-Learning updates. The target network is updated every $C$ time steps as a clone of the current $Q$ network; the use of the target network reduces the chances of oscillations and divergence. A variation of the deep Q-network, known as a Deep Recurrent Q-Network \cite{hausknecht2015deep}, adds a \ac{LSTM} layer to help learn temporal patterns. Additional variations include the double deep Q-network, and \ac{D3QN}. Furthermore, deep reinforcement learning has also been applied to problems with continuous action spaces. In \cite{lillicrap2015continuous}, an actor-critic algorithm known as \ac{DDPG} is presented that is based on the \ac{DPG} algorithm which exploits the idea of experience replay and target networks from the \ac{DQN} as well as batch normalization. \ac{DDPG} is applied successfully to many continuous control problems. In \cite{heess2015memorybased} \ac{RDPG} is introduced as an extension to \ac{DDPG} by the addition of recurrent \ac{LSTM}. 
The characteristics and capabilities of deep reinforcement learning warrant further investigation for its application to autonomous \ac{UAV} applications.

\begin{figure}[h]
\centering
\includegraphics[width=4.5 in]{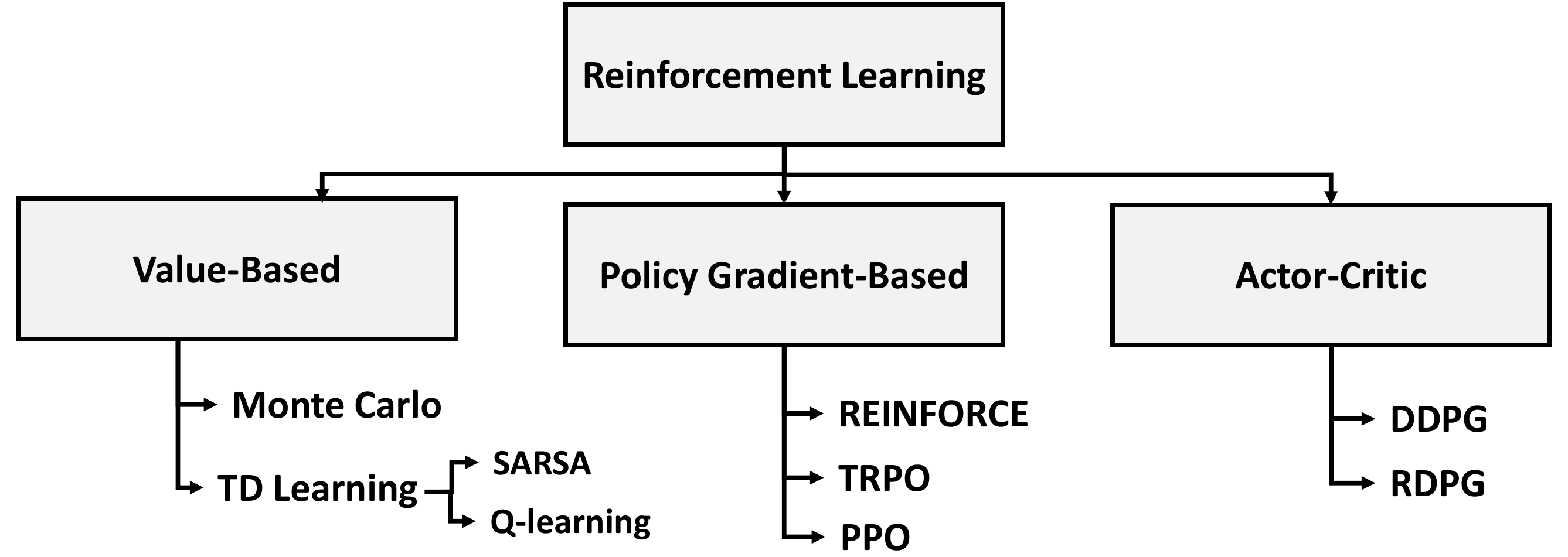}
\caption{Model-free reinforcement learning algorithms}
\label{fig:rloverview}  
\end{figure}

A summary of the different model-free reinforcement learning algorithms is shown in Figure \ref{fig:rloverview}.

\section{Deep Learning for UAS Autonomy}
\label{sec:DeepLearning}

\begin{figure}[h]
\centering
\includegraphics[width=4.6in]{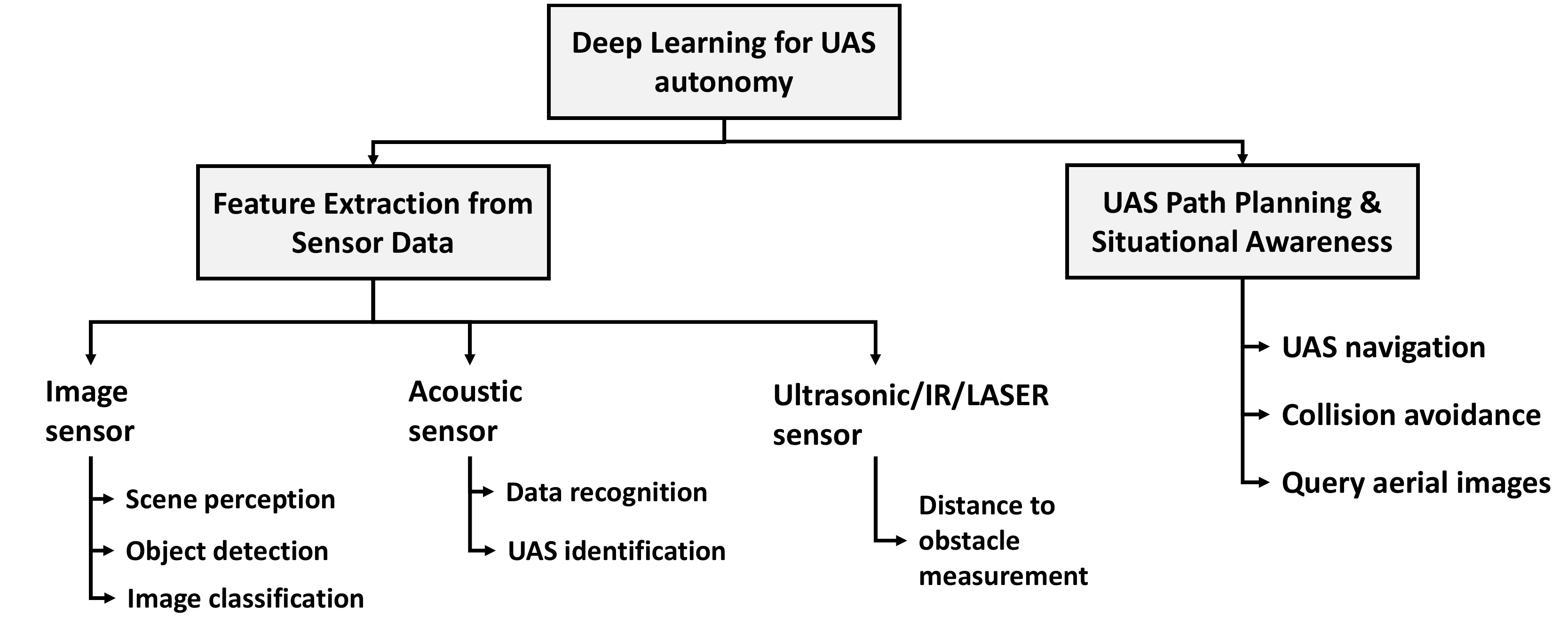}
\caption{Deep learning for UAS autonomy discussed in this section.}
\label{fig:dluas}  
\end{figure}

\emph{Deep learning} has shown great potential in learning complex representations from real environmental data. Its excellent learning capability has shown outstanding results in solving autonomous robotic tasks such as gait analysis, scene perception, navigation, etc., \cite{DL_scene,DL_gait}. The same aspects can be applied for enabling autonomy to the UAS. The various UAS focus areas where deep learning can be applied are scene perception, navigation, obstacle and collision avoidance, swarm operation, and situational awareness. This is also exemplified in the Fig.\ref{fig:dluas}.

Deep learning has been applied as a feature extraction system to learn a high dimensional data representation from the raw sensor output. On the other hand, planning and situational awareness, involve several sub-tasks such as querying or surveying aerial images, navigation control/guidance, collision avoidance, position-dependent control actions, etc. Accordingly, we classify this section into two broad categories: (i) Feature extraction from sensor data and (ii) \ac{UAS} path planning and situational awareness.
\subsection{Feature Extraction from Sensor Data}
\label{sec:fe}
The authors of \cite{Imagery_1} demonstrated the accuracy of a supervised deep learning image classifier to process the monocular images. The classifier predicted outputs of the forest trail direction such as left, right, or straight and claims an accuracy comparable to humans tested on the same image classification task. This scene perception task will require the \ac{MAV} to perceive the trail and react (take actions) to stay on the trail. The authors adopted a typical CNN architecture to accomplish the supervised image classification task. The CNN involved four convolutional layers interlaced with max pooling layers and concluding with two fully connected layers. The output fully connected layer adopted softmax classification that yields the probability of the input image to belong to a particular class. The network was trained using \ac{SGD}. The direction estimates from the CNN were extended to provide navigation control. The navigation control for autonomous trail following was tested on ParrotAR Drone interfaced with a laptop and a standalone quadrotor. The paper reported lower classification accuracy for the real-world testing conditions as opposed to the good quality GoPro images in the training dataset. 


The AlexNet \cite{alexnet} architecture was employed for palm tree detection and counting in \cite{Imagery_3} from aerial images. The images were collected from the QuickBird satellite. A sliding window technique with a window size of $17\times17$ pixels and a stride of 3 pixels was adopted to collect the image dataset. Only a sample with a palm tree located in the center was classified as positive palm tree detection. Spatial coordinates of the detected palm tree classes are obtained and those corresponding to the same palm tree samples are merged. Those spatial coordinates with a Euclidean distance below a certain threshold are grouped into one coordinate. The remaining coordinates represent the actual coordinates of the detected palm trees. The work reported accurate detection of 96\% palm trees in the study area.

\begin{figure}[h]
\centering
\includegraphics[width=4.6in]{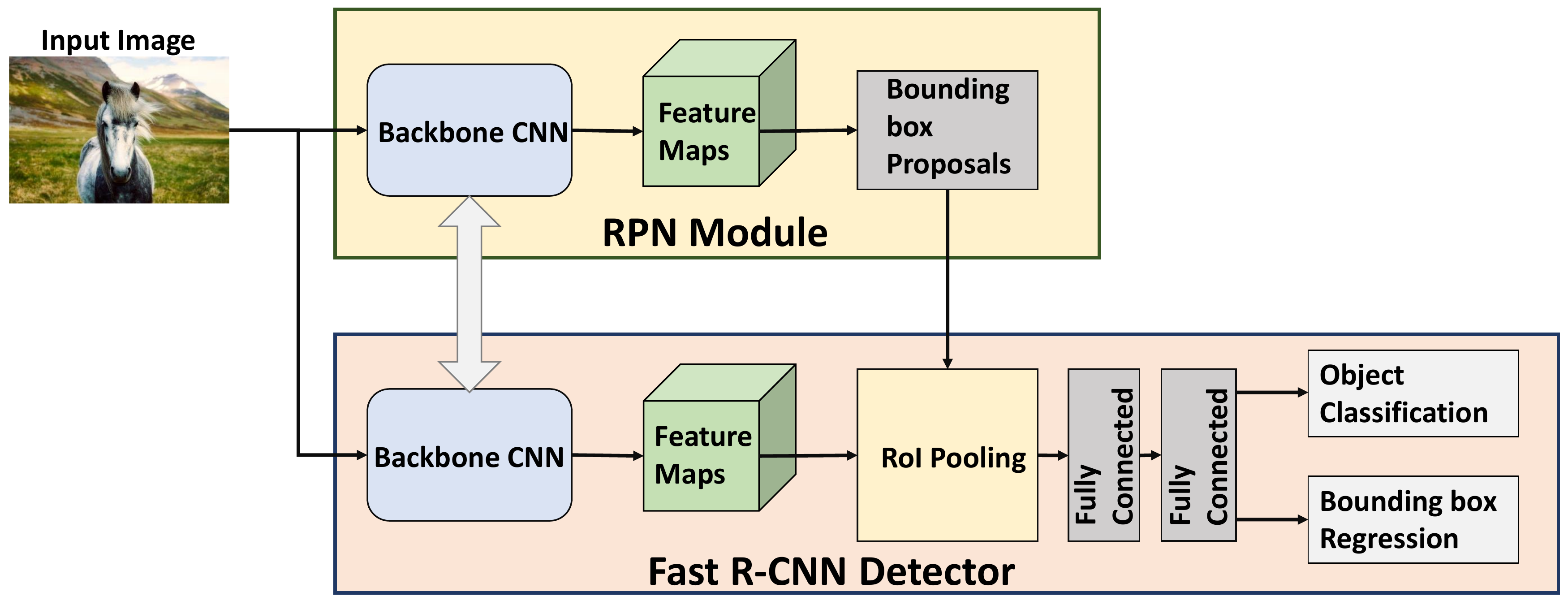}
\caption{Faster R-CNN architecture}
\label{fig:fasterrcnn}  
\end{figure}

Faster R-CNN \cite{fastRCNN} architecture was employed for car detection from low-altitude UAV imagery in \cite{Imagery_4}. Faster R-CNN comprises a region proposal network (RPN) module and a fast R-CNN detector. The RPN module is a deep convolutional architecture that generates region proposals of varying scales and aspect ratios. Region proposals may not necessarily contain the target object. These region proposals are further refined by the fast R-CNN detector. The RPN and fast R-CNN detector modules share their convolutional layers and are jointly trained for object detection. For the car detection task, the VGG-16 model \cite{vgg16} was adopted to form the shared convolutional network. The RPN generates $k$ region proposals in the form of $2k$ box classification and $4k$ box regression outputs. The box regression outputs correspond to the coordinates of the $k$ region proposals while the box classification represents the objectness score, \emph{i.e.,} the probability that each proposal contains the target object (car) or not. The faster R-CNN is trained with a multitask loss function comprising of classification and regression components.
The car detection imagery was collected with GoPro Hero Black Edition-3 mounted on a DJI Phantom-2 quadcopter. The paper reported car detection accuracy of 94.94\% and demonstrated the robustness of the method to scale, orientation, and illumination variations. For a simple exposition, the faster R-CNN architecture is shown in Fig.\ref{fig:fasterrcnn}.
In \cite{Imagery_5}, the faster R-CNN architecture is applied for maize tassel detection from UAV RGB imagery. Here, different CNN architectures were experimented to form the shared layers between the RPN and fast R-CNN detector modules. The paper reported higher accuracy with ResNet \cite{resnet} in contrast to VGGNet for image resolution of $600\times600$ and UAV altitude of 15 m.

The faster R-CNN architecture was compared with You Only Look Once (YOLO v3) \cite{YOLOv3} for car detection from UAV imagery in \cite{Imagery_6}. YOLOv3 is an advancement over its predecessors YOLOv1 \cite{YOLOv1} and YOLOv2 \cite{YOLOv2}. Unlike its predecessors, YOLOv3 can perform multi-label classification of the detected object. Secondly, the bounding box prediction assigns an objectness score of 1 to the predicted box that overlaps the ground truth box more than a predefined threshold. In this way, YOLOv3 assigns one bounding box corresponding to a ground truth object. Additionally, YOLOv3 predicts bounding boxes at 3 different scales. Lastly, it adopts a 53-layered CNN feature extractor named Darknet-53. The study found both YOLOv3 and faster R-CNN performing comparably well in classifying the car object from the image. Although YOLOv3 outperformed faster R-CNN in processing time and sensitivity, \emph{i.e.,} the ability to identify all the cars in the image.

In \cite{Acoustic_1}, a  \ac{PS-DNN} is used for voice identification of people for emergency rescue missions. The microphone array embedded onboard a Parrot Bebop UAV is used for collecting acoustic data. The PS-DNN is posed as a multitask learning framework to achieve two simultaneous tasks - sound source separation and sound source identification. The PS-DNN for multitask learning is a feedforward neural network with partially shared hidden layers between the two sub-networks. Mel filter bank feature vectors obtained by applying windowed \ac{STFT} on the acoustic signals are fed as input to the PS-DNN. The network was trained with Adam learning optimizer \cite{adam} with a learning rate of $2\times10^{-4}$. The study demonstrated promising accuracy when a partially annotated dataset was employed. 

Three \ac{ESC} models - CNN, RNN, and \ac{GMM} - were experimented in \cite{Acoustic_2} to detect commercial drones in real noisy environments. The dataset consisted of ordinary real-life noises and sounds from commercial drones such as 3DR Solo, DJI Phantom-3, DJI Phantom-4, and DJI Inspire. The study demonstrated RNN outperforming the CNN and GMM models. The RNN architecture is a bidirectional \ac{LSTM} with 3 layers and 300 LSTM units. An early-stopping strategy is adopted in the training phase such that if the accuracy and loss do not improve after 3 epochs, the training is stopped. RNN exhibited good generalization over unseen data types with an F-score of 0.6984 and a balanced precision and recall while the CNN resulted in false positives. On the other hand, GMM exhibited better detection performance to CNN but low F-scores deterring practical use.

Drone identification based on acoustic fingerprints using \ac{CNN}, \ac{RNN}, and \ac{CRNN} is presented in \cite{Acoustic_3}. CRNN \cite{CRNN} exploits the advantages of both CNN and RNN to extract spatio-temporal features. The three different architectures were utilized to extract unique acoustic signatures of the flying drones. The authors collected the drone acoustic dataset by recording the sound produced by the drone's propellers while flying them in an indoor environment. Two types of UAVs from the Parrot family named Bebop and Mambo were utilized in this study. The neural networks classify the audio input as drone and not drone. The work portrayed the CNN outperforming both RNN and CRNN in terms of accuracy, precision, F1-score, and recall while RNN exhibited lesser training time. However, the performance of RNN was very poor on all counts which could be attributed to the short duration audio clips as opposed to long sequential data. CRNN, however, outperformed RNN and exhibited comparable performance to that of CNN with the added benefit of lesser training time. The authors also extended their work to multi-label classification to identify the audio clips as Bebop, Mambo, and Unknown. In this task again, a similar performance trend was observed as with the binary classification.

\subsection{UAS Path Planning and Situational Awareness}
\label{sec:sa}
A CNN-based controller strategy for autonomous indoor UAV navigation is considered in \cite{PSA_1}. The limited precision of \ac{GPS} in the indoor environment and the inability to carry heavy weight sensors render indoor navigation a challenging task. The CNN aims to learn a controller strategy to mimic an expert pilot's navigation decisions. The dataset of seven unique indoor locations was collected with a single forward facing camera onboard a Parrot Bebop Drone. The classifier is trained to return flight commands - Move Left, Move Right, Move Forward, Spin Left, Spin Right, and Stop - by training with manually labeled expert flight commands. The CNN classifier followed the CaffeNet \cite{caffenet} architecture with five convolutional layers and three fully connected layers. The classifier was trained on NVIDIA GTX 970M \ac{GPU} with NVIDIA cuDNN \cite{cudnn}. The trained classifier is tested on a combination of familiar and unseen test environments with different objects, lighting, and geometry. The classifier reported success rates in the range of 60\%-80\% for the test locations implying acceptable robustness in flying autonomously through buildings with different objects and geometry.

An interesting approach to UAV navigation is adopted in \cite{PSA_2} where it is taught to fly by crashing. Here, the authors create a crash dataset by crashing the UAV under different scenarios $11500$ times in addition to non-crash data sampled from the same trajectories. In other words, the drone is allowed to learn not to collide into objects by crashing. The collision data is collected by placing the Parrot AR. Drone 2.0 in a random location which is then allowed to takeoff in a random direction and follow a straight line path until the collision. This way the model is allowed to learn if going straight in a specific direction is good or not. The network architecture adopted the AlexNet \cite{alexnet} pre-trained on ImageNet \cite{imagenet}. The pre-trained weights act as initialization for the network weights rather than randomly initialized weights except for the last fully connected layer. The AlexNet architecture involves five convolutional layers and three fully connected layers. The final layer adopts the binary softmax activation function which classifies the navigational actions for the drone. Given an input image, the network decides whether to go left, right or straight. Experimental demonstrations portrayed the efficacy of this supervised learning approach in avoiding glass walls/doors, corridors, and hallways in contrast to an image depth estimation method.

A regression CNN for indoor navigation is proposed in \cite{regCNN}. Autonomous indoor navigation is enabled by predicting the distance to collision based on the visual input from the monocular camera onboard. The authors adopt a self-supervised approach to collect indoor flight dataset annotated with distance to the nearest obstacle in three different diverging directions. The automated annotation is enabled with the help of three pairs of infrared and ultrasonic sensors mounted on the UAV pointing towards different directions with respect to the camera's field of view. The regression CNN follows a two-stream architecture with the first two layers of the streams similar to that of the AlexNet CNN. The two streams are fused to concatenate the feature maps from the streams followed by processing with a convolutional layer similar to the third convolutional layer of AlexNet. The two subsequent convolutional layers in the single-stream section also adopt the last two convolutional layers of AlexNet except for the classifier unit in AlexNet which is replaced by a single fully-connected regression layer. The training of the regression CNN was performed with \ac{SGD} with momentum in 30 epochs with a mini-batch size of 128. The implementation and training were performed in MATLAB on a desktop server with Intel Xeon E5-2630 processor, 64GB of RAM,  and a GTX1080 \ac{GPU}. The UAV is a Parrot AR-Drone 2.0 with a 720p forward-facing camera onboard. During the experiments, a WiFi connection is established between the UAV and a laptop with an  Intel Core i7-6700HQ, 16GB  of  RAM, and a GTX1070  \ac{GPU} to perform the  CNN  inference and motion planning. The authors compared the performance of the proposed regression CNN against two previously discussed state-of-the-art schemes \cite{PSA_1} and \cite{PSA_2}. Regression CNN demonstrated continuous navigation time without collision 4.6$\times$ and 1.7$\times$ more compared to \cite{PSA_1} and \cite{PSA_2} respectively.

A \ac{MAV}-assisted supervised deep learning approach for ground robot path planning to perform search and rescue operation is proposed in \cite{aerialSearchCNN}. The path planning is executed in three stages. The initial stage involves a human operator flying the MAV in vision-assisted mode to localize a goal location such as a ground robot or a victim. During this initial flight, the camera imagery from the MAV is collected for initial terrain classification. The terrain is mapped to obtain a precise elevation map by monocular 3D reconstruction. The CNN classifier is trained on-the-spot without any \emph{apriori} information. The on-the-spot classifier training involves an operator flying the MAV and labeling a few regions of interest from the live camera imagery. Many training patches are gathered from the few labeled regions by cropping patches that fall on previously labeled areas. The authors of \cite{aerialSearchCNN} also report a spot training time of 10 - 15 min on a CNN. Post training, the patches are classified and projected on to the terrain map. After the goal location is found, the second stage involves an autonomous vision-guided flight to a series of waypoints. The path exploration follows an exhaustive search over the candidate paths in order to effectively reduce the response time. The authors demonstrated the efficacy of their approach via simulation as well as field trials. The MAV for field trials was custom built with onboard \ac{IMU}, quadrotor, downward facing camera, onboard Odroid U3 quad-core computer, and PIXHAWK autopilot software. The ground robot for the experiment was a Bluebotics Absolem which is capable of driving over rough terrain. The field trials with canyon and driveway scenarios demonstrated feasible and efficient path exploration over multiple terrain classes, elevation changes, and untraversable terrain.

Another CNN architecture - whereCNN - was proposed in \cite{ground2aerialCNN} to perform ground to aerial geolocalization. The method aims at mapping a street-view query image to its corresponding location on a city-scale aerial-view image. The CNN architecture for cross-view image matching is inspired by Siamese network \cite{siamese} and is comprised of two identical CNNs to learn a shared deep representation across pairs of street and aerial view images. A contrastive loss function is used as the overall loss to train the whereCNN such that the matched pairs are penalized by their squared Euclidean distance and the mismatched pairs by the squared Euclidean distance to a small margin (for the distance that is smaller than the margin). A smaller margin causes the network to be influenced by harder negatives. The dataset is comprised of 78k pairs of Google street view images along with their corresponding aerial view. The whereCNN was trained for 4 days on an NVIDIA Grid K520 \ac{GPU}. The authors demonstrated that the whereCNN trained without sharing parameters between the siamese network entities generalizes reasonably well on unseen data. The method exhibited cross-view matching accuracy of over 22\% for Charleston, San Diego, and San Francisco.

In Table \ref{tab:dluav}, we summarize the deep learning techniques that enable autonomous UAV applications.
\begin{table*}[!h]
\caption{Deep learning for UAV autonomy}
\centering
\def\arraystretch{1.5}%
\begin{tabular}{|p{3.8 cm}|p{2.6 cm}|p{4cm}|}
\hline
\textbf{Proposed solution}       & \textbf{Architecture}       & \textbf{Application}\\ \hline
Giusti et al. \cite{Imagery_1}         &CNN     &Outdoor UAV navigation \\ \hline
Li et al. \cite{Imagery_3}         &AlexNet        &Palm tree detection\newline and counting \\ \hline
Xu et al. \cite{Imagery_4}         &Faster R-CNN    &Car detection from\newline low-altitude UAV imagery \\ \hline
Liu et al. \cite{Imagery_5}         &Faster R-CNN    &Maize tassel detection \\ \hline
Benjdira et al. \cite{Imagery_6}         &Faster R-CNN, \newline YOLOv3                &Car detection from\newline UAV imagery \\ \hline
Morito et al. \cite{Acoustic_1}         &PS-DNN         &Emergency rescue mission \\ \hline
Jeon et al. \cite{Acoustic_2}         &RNN      &Drone identification \\ \hline
S. Al-Emadi et al. \cite{Acoustic_3}         &CNN, RNN, CRNN      &Drone identification \\ \hline
D. K. Kim and T. Chen \cite{PSA_1}         &CaffeNet      &Indoor UAV navigation \\ \hline
Gandhi et al. \cite{PSA_2}         &AlexNet       &Indoor UAV navigation \\ \hline
A. Kouris and C. Bouganis \cite{regCNN}   &CNN      &Indoor UAV navigation \\ \hline
Delmerico et al. \cite{aerialSearchCNN}         &CNN     &UAV-assisted ground \newline robot navigation \\ \hline
Lin et al. \cite{ground2aerialCNN}         &whereCNN      &Ground to aerial geolocalization \\ \hline
\end{tabular}
\\
\label{tab:dluav}
\end{table*}  

\subsection{Open Problems and Challenges}
In this section \ref{sec:DeepLearning}, we discussed the state of the art deep learning techniques for achieving various \ac{UAS} tasks. Specifically, we discussed how deep learning can be leveraged to accomplish feature extraction from sensor data, planning, and situational awareness. However, there exist several open research challenges on the road to achieving complete autonomy of \ac{UAS} tasks. A few of these are enlisted below:
\begin{enumerate}
    \item \emph{Lack of realistic datasets}: The realistic gap between simulated and actual deployed scenarios poses a severe challenge to the deployed deep learning solutions. The diverse scenarios that can be confronted by a UAV in a realistic setting in terms of the varied obstacles in the traversed path, occluded or visually artifacted targets in an object detection task, the effects caused by the sensors on board, etc., are hard to model in a virtual setting. In addition, generating such a realistic dataset from actual UAVs followed by annotating them is a laborious task.
    \item \emph{Fast deep learning:} Generalizing a supervised deep learning solution to unseen data such as those not represented by the training dataset is an open research challenge. On-the-spot learning implying training of the neural network on-the-fly with limited snapshots of the scenario will prove useful in allowing the model to continue learning new scenarios without forgetting past knowledge. The recently introduced model agnostic meta learning (MAML) \cite{maml} opens door to developing such fast learning techniques.
    \item \emph{Resource-heavy deep learning techniques:} The computational complexity of deep learning architectures is another significant hurdle that poses severe constraints on the latency, weight, flight time, power consumption, and cost. Denser architectures require powerful computational platforms such as \acp{GPU} that are often above the prebuilt onboard computational capacity of the UAVs requiring auxiliary computational units. Such additional computational platforms increase the cost, weight, flight time, and power consumption of the UAVs. 
    \item \emph{Vulnerability to cyberattacks:} Vulnerability of the deployed deep learning techniques to various security attacks is a cause of serious concern. Spoofing attacks, signal jamming, identity forging, among others can disrupt the intended UAV operation leading to asset loss and damage. Integrating adversarial learning techniques to the application-specific deep learning approaches can be one way to tackle such security threats.  
\end{enumerate}

\section{Reinforcement Learning for UAS Autonomy}
\label{sec:RL}

Reinforcement learning provides a learning framework allowing agents to act optimally via sequential interactions with its environment. In comparison to supervised or unsupervised learning, reinforcement learning allows the agent to leverage its own experiences derived from environmental interactions. Additionally, reinforcement learning provides a means to specify goals for the agent by means of a reward and penalty scheme. 
These characteristics of reinforcement learning have led to many research efforts on its application to autonomous \ac{UAS} applications. Reinforcement learning has been primarily applied to lower-level control system tasks that regulate the \ac{UAV}'s velocity, attitude, and navigation as well as other higher-level tasks. 

\subsection{UAS Control System}
Stable control of a \ac{UAS} is a complex task due to nonlinear flight dynamics. Traditional control approaches such as \ac{PID} controllers have been successfully used for \ac{UAS} for attitude and velocity control in stable environments. However, the performance of these controllers can deteriorate in dynamic or harsh environments. The main disadvantages of \ac{PID} control being a constant parameter feedback controller are the control efforts are reactive and the controller does not have apriori knowledge of or the ability to learn about the environment. Techniques from adaptive and robust control can provide insights on designing controllers that can adapt to dynamic environments and operate effectively in the presence of uncertainties. However, a shortcoming of these traditional control techniques is that they typically require a mathematical model of the environmental dynamics and do not explicitly learn from past experiences. Reinforcement learning algorithms present a potential solution to the problem of \ac{UAS} control due to their ability to adapt to unknown environments.

There have been many research efforts focusing on the application of reinforcement learning to control systems on a \ac{UAS} \cite{Waslander_2005,BouAmmar_2010,dosSantos_2012,Zhang_2016_MPC,Hwangbo_2017,Lambert_2019,Koch_2019,Bohn_2019}. Much of the research has been focused on quadrotor \acp{UAV}; however, some of the early works involved autonomous helicopters. Many of the reinforcement learning based control systems discussed in this section are for attitude control of the UAV but some of the works consider trajectory tracking and maneuvering as well. Additionally, several algorithmic approaches have been studied including both online and offline methods operating in conjunction with traditional control algorithms as well as \ac{DRL} based approaches.  

Early works of applying reinforcement learning to \ac{UAV} control problems focused on autonomous helicopters \cite{Bagnell_2001, Kim_nips_2004_autonomous_helicopter, Ng_autonomous_inverted_helicopter_2004, Abbeel_2006}. In these works, data was collected from a human pilot flying a remote control helicopter and the dynamics were learned offline. From the learned dynamics, reinforcement learning algorithms were used to design controllers for various maneuvers including hovering, trajectory tracking, and several advanced maneuvers including inverted hovering, flips, rolls, tunnels, and others from the Academy of Model Aeronautics (AMA) remote control helicopter competition.  

The first work that used reinforcement learning for quadrotor UAV control did so for altitude control \cite{Waslander_2005}. A model-based reinforcement learning algorithm that rewards accurate tracking and good damping performance was utilized to find an optimal control policy. To benchmark with a traditional approach, an integral sliding mode controller was also implemented. Tests conducted on \ac{STARMAC} quadrotors showed both the reinforcement learning and integral sliding mode controllers to have comparable performance, both significantly exceeding that of traditional linear control strategies.

In \cite{BouAmmar_2010}, \ac{FVI} is used to design a velocity control system for a quadrotor UAV. The reinforcement learning FVI controller was compared to a cascaded velocity and attitude controller designed using nonlinear control techniques. The performance of each controller was compared using numerical simulations in MATLAB/SIMULINK. While both controllers produced satisfactory results, the reinforcement learning controller was outperformed in terms of settling time but had a lower percent overshoot. The authors stated that a non-parametric approach to value function estimation, such as the use of a wavelet network, may have resulted in better performance for the reinforcement learning controller. The authors emphasized that an advantage of the reinforcement learning controller is that it does not require any prior mathematical knowledge of quadrotor dynamics to yield satisfactory behavior. 

In \cite{dosSantos_2012}, a Learning Automata reinforcement learning algorithm called \ac{FALA} was used to learn the optimal parameters of nonlinear controllers for trajectory tracking and attitude control. Traditional approaches such as \ac{PID}, sliding mode, and backstepping controllers were used to benchmark against \ac{FALA}. The performance of the controllers was analyzed in simulation under varying non-linear disturbances including wind and ground effects. The reinforcement learning tuned controllers outperformed the mathematically tuned controllers in terms of tracking errors. 

In \cite{Zhang_2016_MPC}, an off-policy method, \ac{MPC}, is used for guided policy search for a deep neural network policy for UAV obstacle avoidance. During training, MPC is used to generate control actions for the UAV using knowledge of the full state, this is used along with the state observations to train the policy network in a supervised learning setting. During testing, only the state observations are available to the policy neural network. Simulations were conducted that demonstrated that the proposed approach was able to successfully generalize to new environments.

In \cite{Hwangbo_2017}, a neural network based policy trained using reinforcement learning is used for trajectory tracking and recovery maneuvers. The authors proposed a new reinforcement learning method that uses deterministic policy optimization using natural gradient descent. Experiments were conducted in both simulation and on a real quadrotor UAV, the Ascending Technologies Hummingbird, that demonstrated the effectiveness of the proposed approach. In simulations, the proposed method outperformed the popular algorithms \ac{TRPO} and \ac{DDPG}. The trajectory tracking test resulted in a small but acceptable steady-state error. Additionally, a recovery test where the quadrotor was manually thrown upside down demonstrated autonomous UAV stabilization. A benefit of the proposed algorithm is low computation time; average time of 6 $\mu$s was reported.  

In \cite{Lambert_2019}, deep \ac{MBRL} is used for low-level control of a quadrotor UAV. Deep \ac{MBRL} is used to learn a forward dynamics model of the quadrotor and then \ac{MPC} is used as a framework for control. The algorithms were evaluated using a Crazyflie nano quadrotor. Stable hovering for 6 seconds using 3 minutes of training data was achieved emphasizing the ability to generate a functional controller with limited data and without assuming any apriori dynamics model.   

In \cite{Koch_2019}, multiple neural network based reinforcement learning algorithms are evaluated for attitude control of UAVs. The algorithms that were evaluated include \ac{DDPG}, \ac{TRPO}, and \ac{PPO}. The reinforcement learning algorithms were compared against \ac{PID} control systems for attitude control of UAVs in a simulation environment. The authors also developed an open-source training environment utilizing OpenAI and was evaluated using the Gazebo simulator. The simulations indicated that the agents trained with PPO outperformed a tuned PID controller in terms of the rise time, overshoot, and average tracking error.

In \cite{Bohn_2019}, \ac{PPO} is applied for attitude control of fixed-wing UAVs. The \ac{PPO} method was chosen largely due to the success reported in \cite{Koch_2019}. The PPO controller was trained in a simulation environment to control the attitude (pitch, roll) and airspeed of the UAV to the specified setpoints. The results showed that the DRL controller was able to generalize well to environments with turbulence. The advantages of the DRL controller were emphasized in the high turbulence scenarios by it outperforming the PID controller in multiple performance metrics including success percentage, rise time, settling time, and percent overshoot.

A DRL robust control algorithm for quadrotor \acp{UAV} is presented in \cite{Wang_2019_DPG_IC}. The algorithm uses \ac{DPG} which is an actor-critic method. Furthermore, similar to classical control design, DPG is augmented with an integral compensator to eliminate steady state errors. Additionally, a two phase learning protocol consisting of an offline and online learning phase is defined for training the model. The offline training is completed using a simplified quadrotor model but the robust generalization capabilities are validated in simulation by changing model parameters and adding disturbances. The capability of the model to learn an improved policy online is demonstrated with faster response time and less overshoot compared to original policy learned offline.

\subsection{Navigation and Higher Level Tasks}

In this section, the use of reinforcement learning for higher level planning tasks such as navigation, obstacle avoidance, and landing maneuvers is studied.

In \cite{Imanberdiyev_2016}, a model-based reinforcement learning algorithm is used as a high level control method for autonomous navigation of quadrotor \acp{UAV} in an unknown environment. A reinforcement learning algorithm called TEXPLORE \cite{texplore_paper} is utilized to perform a targeted exploration of states that are both uncertain in the model and promising for the final policy. This is in contrast to an algorithm such as Q-learning that attempts to exhaustively explore the state space. TEXPLORE uses decision trees and random forests to learn the environmental model. In particular, the decision trees are used to predict the relative state transitions and transition effects. A random forest is used to learn several models of the environment as a single decision tree may learn an inaccurate model. The final model is averaged over the decision trees in the random forest. TEXPLORE then performs its targeted exploration using an algorithm called \ac{UCT}. The authors implement and compare the TEXPLORE algorithm to Q-Learning for a navigation task. The navigation task involves the UAV traveling from a start to an end state under battery constraints i.e., the UAV requires a recharge during the mission in order to make it to the goal. The navigation task is performed in a simulated grid environment implemented using ROS and Gazebo. It is shown that the TEXPLORE algorithm learns effective navigation policies and outperforms the Q-Learning algorithm considerably. 

In \cite{pham2018autonomous}, a \ac{PID} and Q-Learning algorithm for navigation of a \ac{UAV} in an unknown environment is presented. The problem is modeled as a finite \ac{MDP}. The environment is modeled as a finite set of spheres with the centers forming a grid, the state of the \ac{UAV} is its approximate position i.e. one of the points on the grid, and the actions available to the agent are head North, South, East, or West. In this work, a constant flight altitude is assumed and thus the state space is two dimensional. The objective of the agent is to navigate to a goal position following the shortest path in an unknown environment. A \ac{PID} and Q-Learning algorithm are used in conjunction to navigate the \ac{UAV} to the goal position in the unknown environment. The Q-Learning algorithm and $\epsilon$-greedy policy are used by the agent to select the next action given the current state. The action is then translated to a desired position and is inputted to the \ac{PID} controller which outputs control commands to the \ac{UAV} to complete the desired action. The proposed algorithm was implemented and tested in both simulation and on a Parrot AR Drone 2.0. In both simulation and experimentation, the \ac{UAV} was able to learn the shortest path to the goal after 38 episodes.

In \cite{pham2018autonomous_functionapprox}, the authors of \cite{pham2018autonomous} utilize an approximated Q-Learning algorithm that employs function approximation in conjunction with the previously described \ac{PID} and Q-Learning control algorithm for \ac{UAV} navigation tasks. Function approximation is used to handle the large state space and to provide faster convergence time. Fixed sparse representation is used to represent the Q table as a parameter vector. Compared to the work in \cite{pham2018autonomous}, the state representation consists of the relative distance of the \ac{UAV} to the goal and relative distances to obstacles in four directions obtained using on-board radar. Both simulation and real tests demonstrated faster convergence and \ac{UAV} navigation to the goal position.

In \cite{Wang_2019_Navigation}, the authors introduce a DRL algorithm, a variant of \ac{RDPG} called Fast-RDPG, for autonomous UAV navigation in large complex environments. The Fast-RDPG differs from RDPG as it uses non-sparse rewards allowing for the agent to learn online and speed up the convergence rate. The reward function design is discussed which includes transition (i.e., progress towards the goal), obstacle proximity penalty, free space, and time step penalty. The Fast-RDPG algorithm outperforms RDPG and DDPG in terms of rate of success, crash, and stray metrics. Generalization of the Fast-RDPG algorithm to environments of different sizes, different target altitudes, and 3D navigation is discussed as well. 

In \cite{singla_drl_oa_2019}, a Deep Recurrent Q-Network with temporal attention is proposed as a \ac{UAV} controller for obstacle avoidance tasks. The model uses a conditional generative adversarial network to predict a depth map from monocular RGB images. The predicted depth map is then used to select the optimal control action. The temporal attention mechanism is used to weight the importance of a sequence of observations over time which is important for obstacle avoidance tasks. The performance of the proposed approach was compared to Deep Q-Network, \ac{D3QN}, and Deep Recurrent Q-Network without temporal attention algorithms and showed superior performance in simulations.

In \cite{Ramos_2018}, a \ac{DRL} algorithm called Deep \ac{DPG} is used to enable an advanced autonomous UAV maneuvering and landing on a moving platform. The authors integrate the Deep \ac{DPG} algorithm into their reinforcement learning simulation framework implemented using Gazebo and \ac{ROS}. The training phase of the proposed approach was conducted in simulation and the testing phases were conducted in both simulation and real flight. The experiments demonstrated the feasibility of the proposed algorithm in completing the autonomous landing task. Additionally, this work showed that agents trained in the simulation are capable of performing effectively in real flights. 

In \cite{uav_auto_land_drl_polvara_2018}, a \ac{DRL} based approach to perform autonomous landing maneuver is presented. The approach relies on a single downward facing camera as the sole sensor. The landing maneuver is considered as a three phase problem: landmark detection, descent maneuver, and touchdown. A hierarchy of two independent \acp{DQN} is proposed as a solution for the landmark detection and descent maneuver problems. The touchdown maneuver is not considered in the research; however, the authors indicated that it may be solved using a closed loop \ac{PID} controller. A \ac{DQN} is employed for the landmark detection component and a double \ac{DQN} is used for the descent. Additionally, the authors propose a new form of prioritized experience replay called \emph{partitioned buffer replay} to handle sparse rewards. Various simulations were conducted that indicated that the proposed \ac{DRL} approach was capable of performing the landing maneuver and could effectively generalize to new scenarios.

In Table \ref{tab:rluav}, we summarize the reinforcement learning techniques that enable autonomous UAV applications.

\begin{table*}[!h]
\caption{Reinforcement for UAS autonomy}
\centering
\def\arraystretch{1.5}%
\begin{tabular}{|p{3.4 cm}|p{3.4 cm}|p{4 cm}|}
\hline
\textbf{Proposed Solution}       & \textbf{Reinforcement Learning Technique}       & \textbf{Application}\\ \hline
J. A. Bagnell and J. G. Schneider \cite{Bagnell_2001}         &Model-based, PEGASUS     &Helicopter control \\ \hline
Kim et al. \cite{Kim_nips_2004_autonomous_helicopter}         &Model-based, PEGASUS     &Helicopter hovering and maneuvers \\ \hline
Ng. et. al.\cite{Ng_autonomous_inverted_helicopter_2004}      &Model-based, PEGASUS     &Helicopter inverted hovering \\ \hline
Abbeel et. al. \cite{Abbeel_2006}                             &Differential Dynamic Programming &Helicopter aerobatic maneuvers \\ \hline
S. L. Waslander and G. Hoffmann \cite{Waslander_2005}         &Model-based; \newline \ac{LWLR},\newline Policy Iteration &Quadrotor altitude control \\ \hline
Bou-Ammar et. al. \cite{BouAmmar_2010}                        &Fitted Value Iteration   &Quadrotor velocity control \\ \hline
S. R. B. dos Santoes et. al. \cite{dosSantos_2012}            &Finite Action-set Learning Automata &Quadrotor trajectory tracking and attitude control \\ \hline
Zhang et. al. \cite{Zhang_2016_MPC}                           &MPC Guided Policy Search &Quadrotor obstacle avoidance \\ \hline
Hwangbo et. al. \cite{Hwangbo_2017}                           &Neural network policy    &Waypoint tracking and recovery tests \\ \hline
Lambert et. al. \cite{Lambert_2019}                           &Deep model-based         &Hovering \\ \hline
Koch et. al. \cite{Koch_2019}                                 &DDPG, TRPO, PPO          &Attitude control \\ \hline
Bøhn et. al. \cite{Bohn_2019}                                 &PPO                      &Attitude control \\ \hline
Y. Wang et. al. \cite{Wang_2019_DPG_IC}                       &DPG                      &UAV control \\ \hline         
Imanberdiyev et. al. \cite{Imanberdiyev_2016}                 &Model-based,TEXPLORE     &UAV navigation \\ \hline
Pham et. al. \cite{pham2018autonomous}                        &Q-Learning               &UAV navigation \\ \hline
Pham et. al. \cite{pham2018autonomous_functionapprox}         &Q-Learning with function approximation &UAV navigation \\ \hline
C. Wang et. al. \cite{Wang_2019_Navigation}                   &Fast-RDPG                &UAV navigation \\ \hline
Singla et. al. \cite{singla_drl_oa_2019}                      &Deep recurrent Q network with temporal attention &Obstacle avoidance \\ \hline
A. Rodriguez-Ramos et. al. \cite{Ramos_2018}                  &DDPG                     &Landing on a moving platform \\ \hline
Polvara et. al. \cite{uav_auto_land_drl_polvara_2018}         &DQN                      &Autonomous landing \\ \hline
\end{tabular}
\\
\label{tab:rluav}
\end{table*}  

\subsection{Open Problems and Challenges}

There are still several open problems and challenges associated with reinforcement learning based autonomous UAV solutions. Many problems and challenges are associated with the transition from simulation to hardware. This is evidenced by limited results on the performance of reinforcement learning solutions performing high complexity planning tasks in real life tests. A challenge associated with the transition is managing the reality gap between simulation and real life testing. Additionally, as deep reinforcement learning solutions are utilized for autonomy, the integration onto an embedded UAV platform can become challenging due to the computational requirements of the algorithms and the \ac{SWaP} constraints of the UAV. 

Other challenge areas include developing algorithmic solutions that enable higher degrees of autonomy. For example, more complex tasks and missions may require the UAV to cooperate with other autonomous systems and/or humans via \acp{NUI}.  Also, the majority of the published works consider scenarios with a static mission objective in dynamic environments; however, in general, the autonomous agent will need to be able to operate in scenarios where both mission objectives and the environment are dynamic. It is also possible that the mission will consist of multiple objectives that need to be completed simultaneously.
 
\section{Simulation Platforms for UAS}
\label{sec:Simulation}

The ability to accurately simulate \ac{UAS} in realistic operational environments is an invaluable capability. This is largely due to the fact that real hardware-based testing of \ac{UAS} is both a time consuming and expensive process. The potential for injuries and damages or losses are the main challenges associated with hardware-based testing. Additional challenges and constraints include limited battery life and the laws and regulations of outdoor flight. These challenges are exacerbated in the context of deep learning and reinforcement learning based autonomy solutions as they require large amounts of training data and experiences in order to learn effective behaviors and are also often unstable during their training phases. Additionally, it can also be challenging and/or costly to collect ample training data for machine learning based autonomous \ac{UAS} algorithms. Physically and visually realistic \ac{UAS} simulations are potential solutions to several of these challenges. For example, a realistic visual simulation of an operational environment could be used to create a dataset for a deep learning algorithm. Furthermore, simulation provides a means to test \ac{UAS} in scenarios that can be hard to create in real life e.g. failure modes, harsh environmental conditions, etc. Simulation also provides a means for establishing easily repeatable environments for algorithm comparisons and software regression testing. 

\subsection{Simulation Suites}

This section now presents a survey of popular simulation software platforms for \ac{UAS}. Previous surveys conducted in \cite{UAV_flight_controller_simulator_survey, Simulation_Hentati, Mairaj_2019} introduced the majority of available \ac{UAS} simulation platforms for various applications. The discussion in this section focuses primarily on open-source simulators that appear useful for research and development of autonomous \ac{UAS} applications.

\begin{figure}[h]
\centering
\hspace{-1 cm}
\includegraphics[width=3 in]{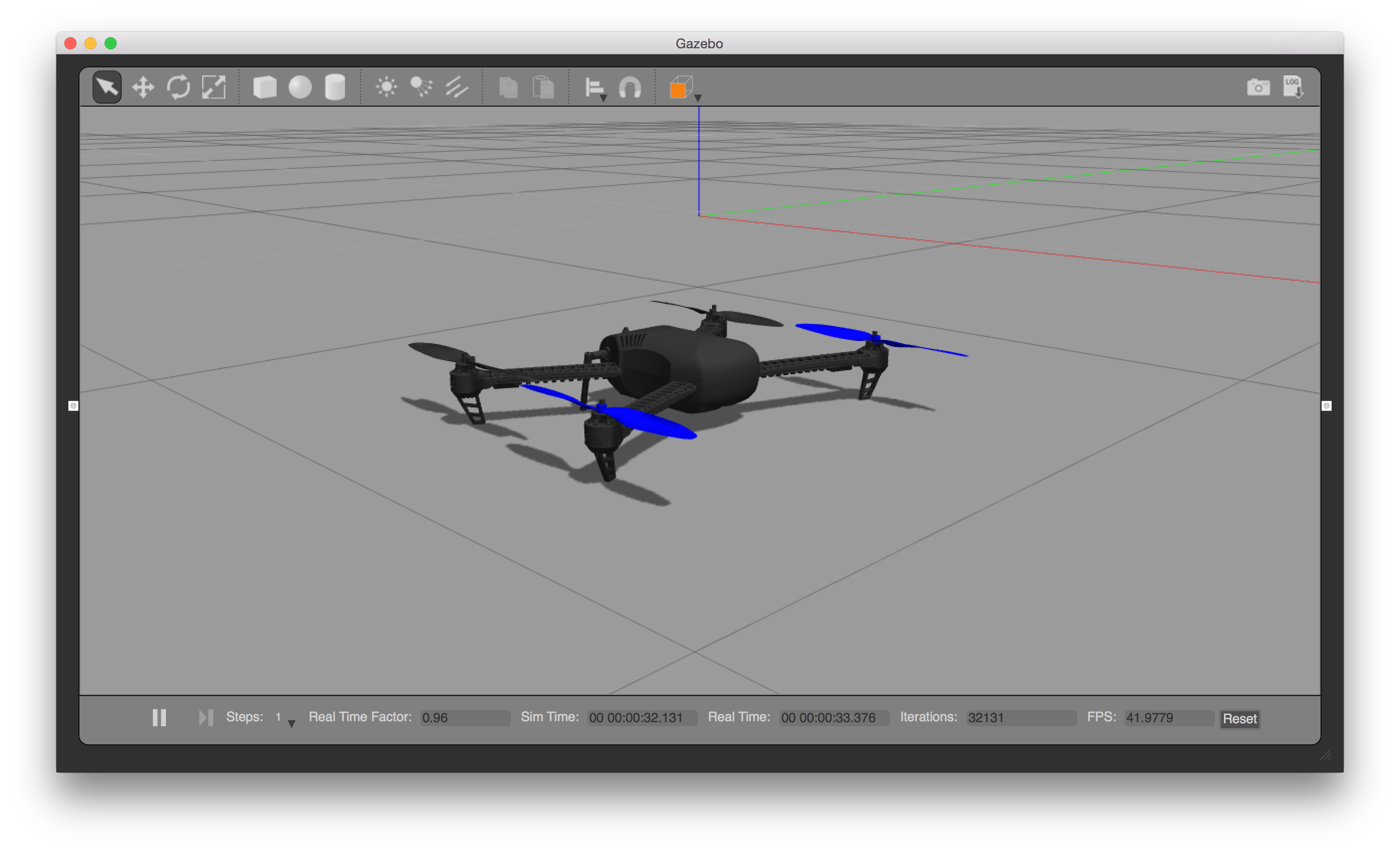}
\caption{UAV simulation in Gazebo \cite{px4_gazebo_pic}}
\label{fig:gazebo}  
\end{figure}

Gazebo \cite{Gazebo_website, Gazebo_paper} is an open-source robotics simulator capable of simulating multiple robots in both indoor and outdoor environments. This is enabled by its integration with high-performance physics engines, e.g., \ac{ODE}, Bullet, Simbody, and \ac{DART} as well as its ability to model various sensors, noise, and environmental effects. The Gazebo architecture is modular by allowing for worlds and objects to be defined using \ac{SDF} files while enabling sensor and environmental effect modules to be added as plugins. \ac{OGRE} \cite{Ogre_website} is utilized by Gazebo for high fidelity visual rendering of the environment that captures different textures and lighting. Gazebo is also one of the default simulators integrated with the popular robotics middleware package \ac{ROS}. By itself, Gazebo does not provide the capability to simulate \ac{UAV}s; however, there have been multiple works that define the necessary model, sensor, and controller plugins to facilitate \ac{UAV} simulation and is discussed herein. An example of \ac{UAV} simulation using Gazebo is shown in Figure \ref{fig:gazebo}.

In \cite{hector_quadrotor_paper,hector_quadrotor_wiki} simulation of quadrotor \acp{UAV} using Gazebo and \ac{ROS} is implemented as an open-source package called hector\_quadrotor. The hector\_quadrotor package provides the geometry, dynamics, and sensor models for quadrotor \acp{UAV}. Sensor models for \ac{IMU}, barometer, ultrasonic sensor, magnetic field, and \ac{GPS} in addition to the default sensor models provided by Gazebo such as LIDAR and cameras. \acp{EKF} and cascaded \ac{PID} controllers are implemented and utilized for state estimation and control respectively. A tutorial example of integrating a LIDAR based \ac{SLAM} algorithm with the simulated \acp{UAV} is included in the package's documentation.

RotorS is another open-source \ac{MAV} simulator using Gazebo and \ac{ROS} \cite{RotorS_chapter,rotorS_wiki}. Models of various multirotor \acp{UAV} including the AscTec Hummingbird, AscTec Pelican, and AscTec Firefly are included with the simulator. Default simulator sensors include \ac{IMU}, a generic odometry sensor, and visual inertial sensor. Similar to hector\_quadrotor, RotorS provides a baseline UAV simulation using Gazebo by defining the required UAV, sensor, and controller configuration files and plugins. The RotorS package provides a well documented and functional UAV simulator that a researcher can use for rapid prototyping of new autonomous UAV control algorithms.

In \cite{Koch_2019}, a framework called GymFC for tuning \ac{UAV} flight control systems was introduced. The framework integrates the popular reinforcement learning toolkit OpenAI Gym \cite{OpenAI_Gym} and the Gazebo simulator to facilitate research and development of attitude flight control systems using \ac{DRL}. GymFC defines three layers to provide seamless integration of reinforcement learning based \ac{UAV} control algorithms: Digital Twin Layer, Communication Layer, and Environment Interface Layer. The Digital Twin Layer consists of the simulated \ac{UAV} and environment as well as interfaces to the Communication Layer. The Communication Layer is the interface between the Digital Twin and Environment Interface Layer that implements lower level functionality to enable control of the \ac{UAV} and the simulation. The Environment Interface Layer implements the environmental interface defined by the OpenAI Gym API that the reinforcement learning agent interacts with. In the original work \cite{Koch_2019}, the proposed \ac{DRL} based attitude controllers were only evaluated in simulation. The open-source Neuroflight framework \cite{koch2019neuroflight} has since been introduced for deploying neural network based low-level flight control firmware on real \acp{UAV}. Neuroflight utilizes GymFC for initial training and testing of controllers in a simulation environment and then deploys the trained models to the \ac{UAV} platform. Initial tests of Neuroflight have demonstrated stable flight and maneuver execution while the neural network based controller runs on an embedded processor onboard the \ac{UAV}.

The Aerostack software framework \cite{Aerostack_Paper, aerostack_git, Simulation_Sanchez-Lopez} defines an architectural design to enable advanced \ac{UAV} autonomy. Additionally, Aerostack has been used for autonomous \ac{UAV} research and development in both simulations (utilizing the RotorS simulator \cite{RotorS_chapter}) and in hardware such as the Parrot AR Drone.

\begin{figure}[h]
\centering
\hspace{-1 cm}
\includegraphics[width=3 in]{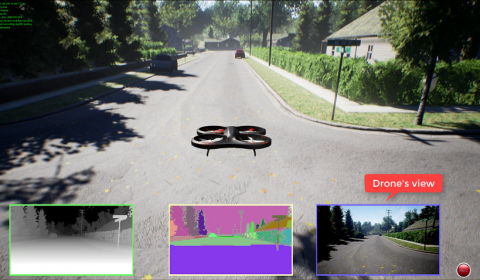}
\caption{UAV simulation in AirSim \cite{airsim_pic}}
\label{fig:airsim}  
\end{figure}

Microsoft AirSim \cite{airsim2017fsr, airsim_git} is an open-source simulator for both aerial and ground vehicles. AirSim provides realistic visual rendering of simulated environments using the Unreal Engine, as shown in Figure \ref{fig:airsim}. AirSim was designed as a simulation platform to facilitate research and development of \ac{AI} enabled autonomous ground and aerial vehicles which motivates its use when developing deep learning and reinforcement learning \ac{UAS} solutions. The software is cross platform and can be used on Linux, Windows, and Macintosh operating systems. The AirSim software comes with extensive documentation, tutorials, and \acp{API} for interfacing with vehicles, sensors, and environment for programmatic control and data collection for model training. Recently, AirSim was used as a platform to host a simulation-based drone racing competition called Game of Drones \cite{madaan2020airsim}. 

A final consideration is that the popular flight control stacks - PX4 and ArduPilot (discussed in detail in section \ref{sec:flightstack}) - can both integrate with Gazebo and AirSim for software-in-the-loop and hardware-in-the-loop simulations. The Gazebo interfaces maintained by PX4 are derived from the RotorS project.

\subsection{Open Problems and Challenges}

Even with the advances made in the realm of UAS simulations, there are still multiple problems and challenges associated with it. The first problem is typical to any open-source platforms used in different domains - there is no official or industry accepted standard platform. For example, the two most popular open-source flight control stacks, ArduPilot and PX4, both support multiple simulators but there is not a specified official/default simulator common to them. At this time, it appears that both Gazebo or AirSim have the potential for use in autonomous \ac{UAS} research and development. A challenge associated with the Gazebo simulator is that although it is widely used in \ac{UAS} simulation, it technically does not provide native \ac{UAS} simulation support. Works such as \cite{hector_quadrotor_paper, RotorS_chapter} implement the required plugins, configuration, and baseline controllers to enable \ac{UAV} simulation using Gazebo. Additionally, as common with open-source software, there is often limited software maintenance, development support, and documentation of the open-source simulators.

An additional challenge associated with \ac{UAS} simulation is that there can be steep learning curves associated with advanced usage and software development. It appears to be straightforward to install and run examples provided by the simulator; however, it may take time to familiarize with simulator configurations, development workflow, and software APIs. For example, a developer may be required to add support for a new \ac{UAV} platform, sensor type, or environment tailored for the research application. This problem could be mitigated to an extent as the use of these platforms become widespread and if there is a uniform standard to add new features that can be made available to the community. 

An open problem is assessing the reality gap between simulation and real life deployment. This problem will be further studied as research and development of algorithms for autonomous \ac{UAS} continues. Other open problems are associated with the seemingly limited consideration of \ac{UAV} swarm operation, human interaction via \acp{NUI} or ground control stations, and communication systems utilized by the UAS. 

\section{UAV Hardware for Rapid Prototyping}

Rapid \ac{UAS} hardware-based prototyping is an essential step in deploying and validating machine learning solutions. Certain factors such as the unique requirements of the deep learning solution and the cost of \ac{COTS} UAS in commercial market are the driving factors in choosing the custom prototyping route. The requirements of deep learning solutions could be unique to the problem under consideration and consequently the needs would vary. For instance, an object detection task might require a stable flight platform with good quality image sensor. However, a target tracking or acoustic-based search and rescue mission might require maneuverable platform with image sensor and acoustic sensors onboard respectively. UAS prototyping for testing deep learning solutions involve several steps such as choosing the appropriate hardware platform, sensors, computational resources, memory unit, flight controller software, among others which depend on the size, weight, and onboard carrying capacity of the UAS platform. This section will serve as a comprehensive guide in choosing the appropriate UAS platform, flight stack software, computational resources as well as the various challenges incurred in UAS prototyping.

\subsection{Classification Choice}
\label{sec:uasclass}
\acp{UAV} are classified based on their wings, size, landing, etc., as seen in the beginning of the chapter (section \ref{subsec:classUAS}). In this section, however, we will focus on the Fixed-wing and Rotary-wing \acp{UAV}. The various UAV classifications will guide the reader in understanding the nuances of the platforms in terms of its hovering, maneuvering, payload capabilities, among others allowing application-specific selection.

\begin{figure*}[h]
\minipage[b]{0.5\textwidth}
\centering
\includegraphics[width=2.3 in]{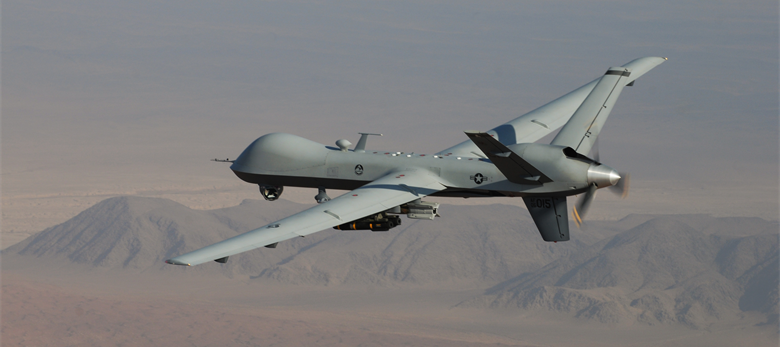}
\caption{Fixed Wing \ac{UAV} \cite{airForce_2015} }
\label{fig:fixed_wing}  
\endminipage\hfill
\minipage[b]{0.49\textwidth}
\centering
\includegraphics[width=2.3 in]{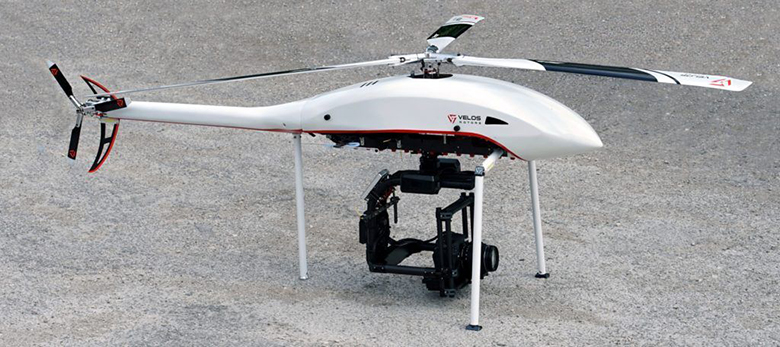}
\caption{Single-Rotor \ac{UAV} \cite{shan2018} }
\label{fig:single_rotor}  
\endminipage\hfill
\end{figure*}

Fixed-wing UAV has rigid wings with airfoil allowing it to produce the desired lift and aerodynamics by deflecting the oncoming air. Although they cannot hover at a place and maintain low speed, they support long endurance flights. Further, they require an obstruction-free runway to take-off and land. However, in comparison to rotary-wing, they carry heavier payloads and are energy-efficient owing to their gliding characteristic. The MQ-9 Reaper is an example of a fixed-wing UAV as in Fig. \ref{fig:fixed_wing}.

Rotary-wing UAV possesses two or more rotary blades positioned around a fixed mast to achieve the desired aerodynamic thrust. Rotary-wing platforms are capable of hovering tasks, low-altitude flights, and perform \ac{VTOL}. In contrast to fixed-wing, they present flexible maneuverability advantages owing to the rotary blades. Rotary-wing \acp{UAV} are further classified into single-rotor, multi-rotor, and fixed-wing hybrid \cite{chapman_2019}.

Single-rotor \acp{UAV} rely on a single front rotor to stay airborne. Although, they possess a tail rotor to control the heading as in Fig. \ref{fig:single_rotor}, it does not count towards the rotor count. The required airflow to move forward is generated by the rotor blades. They are also capable of \ac{VTOL} and hovering tasks. Since they rely on a singular rotor to stay elevated, the blades are usually longer. In contrast to multi-rotor \acp{UAV}, they can carry heavier payloads and are energy-efficient owing to lesser power requirements for a single rotor. The energy-efficient operation enables longer flight times when compared to multi-rotor platforms. Therefore, single-rotor platforms might present themselves as beneficial for aerial surveying applications which require carrying heavier payloads and extended flight times. Helicopters are an example of single-rotor \acp{UAV}.
\begin{figure*}[h]
\minipage[b]{0.5\textwidth}
\centering
\hspace{-1 cm}
\includegraphics[width=1.95 in]{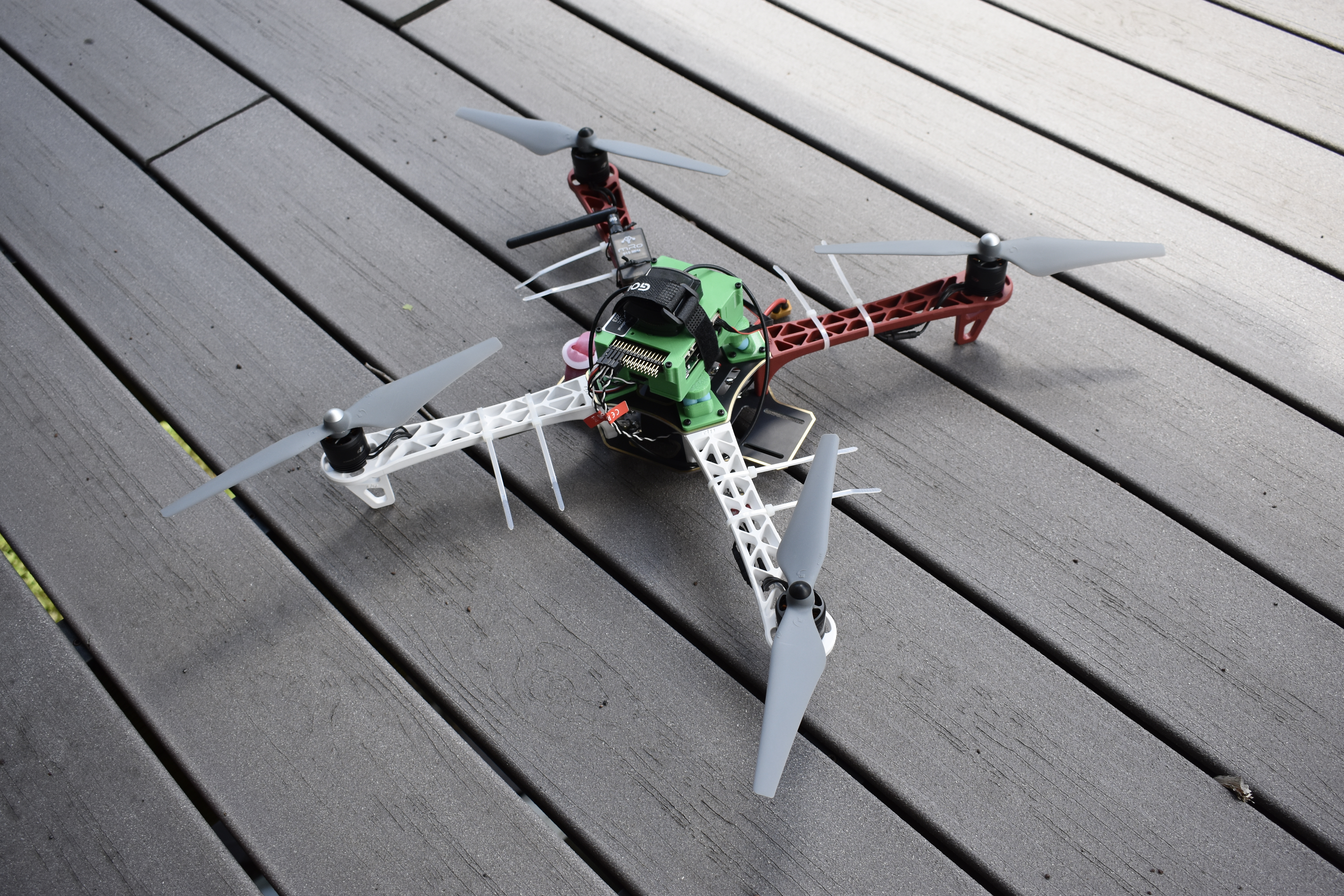}
\caption{Multi-Rotor \ac{UAV}}
\label{fig:multi_rotor}    
\endminipage\hfill
\minipage[b]{0.49\textwidth}\hspace{-0.5cm}
\centering
\includegraphics[width=2.4 in]{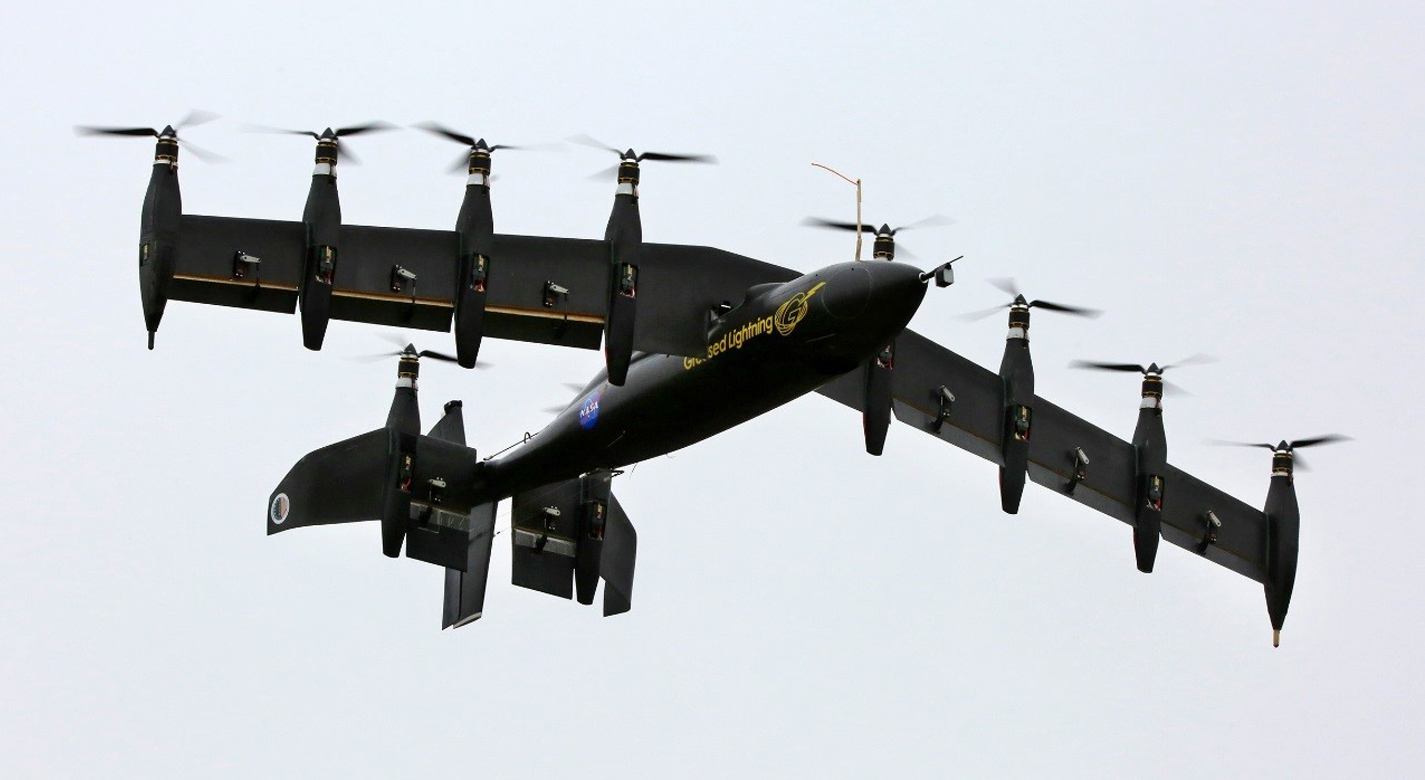}
\caption{Fixed Wing Hybrid \ac{UAV} \cite{nasa} }
\label{fig:fixed_wing_hybrid} 
\endminipage\hfill
\end{figure*}

Multi-rotor \acp{UAV}, on the other hand, uses multiple rotor blades to achieve the desired aerodynamic thrust for lifting and propelling as in Fig. \ref{fig:multi_rotor}. Most common examples of this category are tricopter, quadcopter (quadrotor), hexacopter, and octocopter. Multi-rotor platforms can perform complex maneuvering and hovering tasks but have limited payload capability and flight endurance. They also provide a stable platform for aerial inspection, photography, and precision agriculture applications.

Fixed-wing hybrid UAV platforms combine the aerodynamic benefits of fixed-wing and rotary-wing UAV classes (Fig.\ref{fig:fixed_wing_hybrid}). This coupling adds the \ac{VTOL}, hovering, increased flight speed, and long endurance capabilities. Owing to the fairly recent arrival of the hybrid class, there are still very few developmental resources available for this class. The discussion in this section will enable the developer in choosing the appropriate UAV platform tailored to meet the requirements pertinent to their unique machine learning solution.

\subsubsection*{Build or Buy}
Here, we will ponder upon the pros and cons of buying versus building a UAV. Commercial \acp{UAV} available in the market would serve as an easier and cost-friendly option to rapidly test deep learning solutions. However, specific mission requirements might urge towards building a custom model. 

Commercial \acp{UAV} are often preprogrammed and tested for stability. Most of them come in a ready-to-fly state requiring minimal setup out of the box. The prebuilt \acp{UAV} offer limited customization and could be difficult to repair and/or replace components. An essential requirement for deep learning solutions is the computational power, however, prebuilt UAV platforms have limited onboard computational resources requiring external processors. A costlier option could be purpose-built commercial \acp{UAV} with custom attachments to fit the mission requirements.

UAV prototyping, on the contrary, offers several benefits. Often developers can add custom sensors, batteries, and computational units to a flight-ready UAV platform for rapid deployment and testing. The lift and payload capacity of the UAV judges its flight endurance and stability. Achieving flight stability is guided by several factors such as the right component balance and the ground controller's pilot skills. Building a flight-ready UAV would entail requiring immense electrical and mechanical skills which could be envisioned as a pro as well as a con. The prototyping procedure could be time-consuming while garnering the electro-mechanical skills would be knowledgeable. Another major requirement while building custom prototypes would be the flight controller software needed to control and navigate the \acp{UAV}. To conclude, we have listed a few commercial drones and their specifications in Table \ref{table:drones}. The next subsection sheds light on the flight stack software. 

\begin{table}[h!]
\caption{Commercially available drones}
\centering
\def\arraystretch{1.4}%
\resizebox{\textwidth}{!}{%
\begin{tabular}{|l|c|c|c|c|}
\hline
\multicolumn{1}{|c|}{{\textbf{UAV platform}}} &
  \textbf{Specifications} &
  \begin{tabular}[c]{@{}c@{}}\textbf{Onboard/}\\ \textbf{External} \\\textbf{DL Processing}\end{tabular} &
  \textbf{SDK} &
  \textbf{Estimated Cost} \\ \hline
Ryze Tello EDU \cite{ryze_tello} &
  \begin{tabular}[c]{@{}c@{}}87 g Weight, 13min Flight, \\ WiFi 802.11n, Range Finder, Barometer LED, Camera\end{tabular} &
  External via SDK &
  Tello-Python &
  \$129.00 \\ \hline
DJI Inspire 2 \cite{dji_inspire_2} &
  \begin{tabular}[c]{@{}c@{}}3.44 kg Weight, 4.25 kg payload, 27 min fight time, \\ 2.4000 GHz-2.4835GHz, 5.725 GHz-5.850GHz, GPS, GLONASS, \\ GALILEO, Camera, Vision systems for obstacle avoidance\end{tabular} &
  External via SDK &
  Mobile SDK &
  \$3,299.00 \\ \hline
DJI Matrice 100 \cite{dji_matrice_100} &
  \begin{tabular}[c]{@{}c@{}}2.355 kg weight, 3.6 kg payload, 13 - 40 min flight time, \\ 5.725-5.825 GHz, 922.7MHz-927.7 MHz, \\ 2.400-2.483 GHz (Lightbridge)\end{tabular} &
  \begin{tabular}[c]{@{}c@{}}Onboard via \\ Manifold 2-C \\ or Manifold 2-G\end{tabular} &
  \begin{tabular}[c]{@{}c@{}}Onboard SDK, \\ Mobile SDK\end{tabular} &
  N/A \\ \hline
\multicolumn{1}{|l|}{\begin{tabular}[l]{@{}l@{}}DJI Matrice\\ 200 Series V2 \cite{dji_matrice_200}\end{tabular}}  &
  \begin{tabular}[c]{@{}c@{}}4.91 kg weight, 1.23 kg payload, 33 min flight time, \\ 2.4000-2.4835 GHz,  5.725-5.850 GHz, \\ Different Payload configurations\end{tabular} &
  \begin{tabular}[c]{@{}c@{}}Onboard via \\ Manifold 2-C \\ or Manifold 2-G\end{tabular} &
  \begin{tabular}[c]{@{}c@{}}Onboard SDK\\ Payload SDK\\ Mobile SDK\end{tabular} &
  Request Quote \\ \hline
\multicolumn{1}{|l|}{\begin{tabular}[l]{@{}l@{}}DJI Matrice\\ 300 RTK \cite{dji_matrice_300}\end{tabular}} &
  \begin{tabular}[c]{@{}c@{}}6.3 kg weight, 2.7 kg payload, 55 min flight time, \\2.4000-2.4835 GHz, 5.725-5.850 GHz, \\ Camera Gimbal, infrared ToF Sensing System, FPV Camera, GPS\end{tabular} &
  \begin{tabular}[c]{@{}c@{}}Onboard via \\ Manifold 2-C \\ or Manifold 2-G\end{tabular} &
  \begin{tabular}[c]{@{}c@{}}Onboard SDK\\ Payload SDK\\ Mobile SDK\end{tabular} &
  Request Quote \\ \hline
 \multicolumn{1}{|l|}{\begin{tabular}[l]{@{}l@{}}DJI Matrice\\ 600 Pro \cite{dji_matric_600}\end{tabular}} &
  \begin{tabular}[c]{@{}c@{}}9.5 kg weight, 15.5 kg payload, 16 - 38 min flight time, \\ 920.6 MHz-928 MHz, 5.725 GHz-5.825 GHz, 2.400 GHz-2.483 GHz, \\ Camera Gimbal, Collision avoidance system, GPS, GLONASS\end{tabular} &
  \begin{tabular}[c]{@{}c@{}}Onboard via \\ Manifold 2-C \\ or Manifold 2-G\end{tabular} &
  \begin{tabular}[c]{@{}c@{}}Onboard SDK, \\ Mobile SDK\end{tabular} &
  \$5,699.00 \\ \hline
\multicolumn{1}{|l|}{\begin{tabular}[l]{@{}l@{}}DJI Mavic 2 \\ Enterprise \cite{dji_mavic_2_enterprise}\end{tabular}} &
  \begin{tabular}[c]{@{}c@{}}905 g weight, 1100 g payload, 29 min flight time, \\ 2.400-2.4835 GHz, 5.725-5.850 GHz, GPS, GLONASS, Visual Camera, \\ Omnidirectional Obstacle Sensing, Speaker, Beacon, Spotlight\end{tabular} &
  External via SDK &
  \begin{tabular}[c]{@{}c@{}}Mobile SDK, \\ Windows SDK\end{tabular} &
  Request Quote \\ \hline
 \multicolumn{1}{|l|}{\begin{tabular}[l]{@{}l@{}}DJI Mavic 2 \\ Enterprise Dual \cite{dji_mavic_2_enterprise}\end{tabular}}&
  \begin{tabular}[c]{@{}c@{}}899 g weight, 1100 g payload, 29 min flight time, \\ 2.400-2.4835 GHz, 5.725-5.850 GHz, GPS, GLONASS, \\Thermal Camera, Visual Camera, Camera, Speaker,\\  Omnidirectional Obstacle Sensing, Beacon, Spotlight\end{tabular} &
  External via SDK &
  \begin{tabular}[c]{@{}c@{}}Mobile SDK, \\ Windows SDK\end{tabular} &
  Request Quote \\ \hline
DJI Mavic 2 Pro \cite{dji_mavic_2} &
  \begin{tabular}[c]{@{}c@{}}905 g weight, 31 min flight time, \\ 2.400-2.4835 GHz, 5.725-5.850 GHz, GPS, \\ GLONASS, Pro Camera, Omnidirectional Obstacle Sensing\end{tabular} &
  External via SDK &
  \begin{tabular}[c]{@{}c@{}}Mobile SDK, \\ Windows SDK\end{tabular} &
  \$1,599.00 \\ \hline
DJI Mavic 2 Zoom \cite{dji_mavic_2} &
  \begin{tabular}[c]{@{}c@{}}905 g weight, 31 min flight time, \\ 2.400-2.4835 GHz, 5.725-5.850 GHz, GPS, \\ GLONASS, Zoom Camera, Omnidirectional Obstacle Sensing\end{tabular} &
  External via SDK &
  \begin{tabular}[c]{@{}c@{}}Mobile SDK, \\ Windows SDK\end{tabular} &
  \$1,349.00 \\ \hline
DJI P4 Multispectral \cite{dji_p4} &
  \begin{tabular}[c]{@{}c@{}}1487 g weight, 27 min flight time, \\ 2.4000 GHz-2.4835 GHz, 5.725 GHz-5.850 GHz,\\  GPS, GLONASS, GALILEO, RGB Camera, 5 monochome sensors\end{tabular} &
  External via SDK &
  Mobile SDK &
  Request Quote \\ \hline
 \multicolumn{1}{|l|}{\begin{tabular}[l]{@{}l@{}}DJI Phantom 4 \\ Pro V2.0 \cite{dji_phantom_pro_v2}\end{tabular}}&
  \begin{tabular}[c]{@{}c@{}}1375 g weight, 30 min flight time, \\ 2.4000 GHz-2.4835 GHz, 5.725 GHz-5.850 GHz, \\ GPS, GLONASS, GALILEO, RGB Camera, infrared sensors\end{tabular} &
  External via SDK &
  Mobile SDK &
  \$1,599.00 \\ \hline
\multicolumn{1}{|l|}{\begin{tabular}[l]{@{}l@{}}DJI Phantom 4 \\ RTK \cite{dji_phantom_4_rtk}\end{tabular}}&
  \begin{tabular}[c]{@{}c@{}}1391 g weight, 30 min flight time, \\ 2.4000 GHz-2.4835 GHz, 5.725 GHz-5.850 GHz, \\ GPS, GLONASS, GALILEO, RGB Camera, infrared sensors\end{tabular} &
  External via SDK &
  Mobile SDK &
  Request Quote \\ \hline
\multicolumn{1}{|l|}{\begin{tabular}[l]{@{}l@{}}Parrot ANAFI \\ ANAFI Thermal \cite{parrot_anafi_thermal}\end{tabular}} &
  \begin{tabular}[c]{@{}c@{}}315 g weight, 26 min flight time, \\ Wi-Fi 802.11a/b/g/n, GPS, GLONASS, Barometer, magnetometer,\\ vertical camera, ultra sonar, 6 axis, IMU,3 axis accelerometer, \\3 axis gyroscope, thermal imaging camera, 4k camera\end{tabular} &
  External via SDK &
  \begin{tabular}[c]{@{}c@{}}Parrot \\ Ground SDK\end{tabular} &
  \$1,900.00 \\ \hline
Parrot ANAFI USA \cite{parrot_anafi_usa} &
  \begin{tabular}[c]{@{}c@{}}500 g weight, 32 min flight time, \\ Wi-Fi 802.11a/b/g/n, GPS, GLONASS, GALILEO, Barometer, \\ magnetometer, vertical camera, ultra sonar, 6 axis, IMU, 4k camera \\ 3 axis accelerometer, 3 axis gyroscope, 32x zoom camera\end{tabular} &
  External via SDK &
  \begin{tabular}[c]{@{}c@{}}Parrot \\ Ground SDK\end{tabular} &
  Coming soon \\ \hline
Parrot ANAFI Work \cite{parrot_anafi_work} &
  \begin{tabular}[c]{@{}c@{}}321 g weight, 25 min flight time, \\ Wi-Fi 802.11a/b/g/n, GPS, GLONASS, Barometer, magnetometer, \\ vertical camera, ultrasonar, 6 axis, IMU, 3 axis accelerometer,\\ 3 axis gyroscope, thermal imaging camera, 4k camera\end{tabular} &
  External via SDK &
  \begin{tabular}[c]{@{}c@{}}Parrot \\ Ground SDK\end{tabular} &
  \$999.00 \\ \hline
\end{tabular}%
}
\label{table:drones}
\end{table}

\subsection{Flight Stack}
\label{sec:flightstack}
\begin{figure}[h]
\centering
\includegraphics[width=4.5 in]{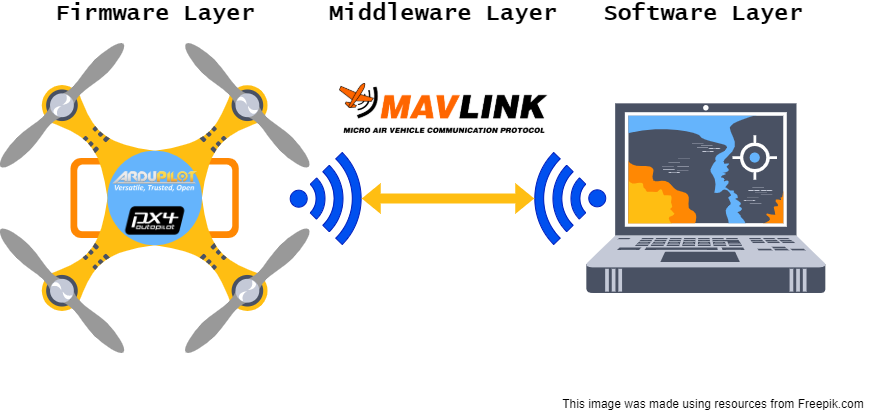}
\caption{Flight Stack}
\label{fig:flight_stack}  
\end{figure}

Flight stack is the flight controller software that comprises of a set of positional, navigational guidance and control algorithms, interfacing, and communication links that directs the UAV's flight path and maneuverability. A flight stack is typically comprised of firmware, middleware, and interface layers as in Fig. \ref{fig:flight_stack} \cite{caleberg_2019} whereby the middleware supports the communication link to enable command and control (C2) and telemetry data message passing. The software layer performs the interfacing of the firmware via the communication link protocol. Software layer refers to the \ac{GCS} software that performs UAV configuration and monitoring.

There are many open-source flight controller software available today namely; ArduPilot, PX4, Paparazzi, among others. Flight controller software enables autonomous operation capability to specific UAV platforms (airframes). This comprises fault detection and handling, C2 link protocol, battery monitoring, obstacle avoidance, landing, return home features, data logging, among others. The fault detection and handling support features such as landing when missing C2 link, return to home when missing C2 link, automatic parachute release, battery voltage warning, geofence, land/return to home when battery low, safety check for sensor error, etc. Some of the C2 link protocols are MavLink, UAVTalk, XBUS, XBee, FrSky, HoTT, \ac{PPM}, and Lightweight TeleMetry (LTM).

ArduPilot is an open-source flight controller software released under GNU General Public License (GPL) which supports a wide range of vehicles including fixed-wing UAV, multi-rotor UAV, single-rotor UAV, boats, and submarines \cite{ardu}. It can be run on a Linux-based operating system (OS) allowing support on single-board computers to full PC systems. ArduPilot has a desktop \ac{GCS} software for mission planning and calibration for Linux, Windows, and Macintosh OS. It also supports MAVLink, FrSky, and LTM communication protocols. ArduPilot additionally supports the usage of multiple radio control receivers for redundancy, failover, and/or handoffs.

PX4 flight controller \cite{px4} from DroneCode collaborative project \cite{dronecode} is released under \ac{BSD} license and supports both fixed-wing and multi-rotor airframes. PX4 enables operation with QGroundControl GCS software from where the UAV can be configured as well as monitored. Both ArduPilot and PX4 supports satellite waypoint navigation and satellite position hold. ArduPilot and PX4 additionally support stereovision navigation function and follow me autonomous navigation features respectively. 

Paparazzi flight controller supports fixed-wing, flapping-wing, hybrid, and multi-rotor airframes and is released for public use under GNU GPL \cite{paparazzi}. The GCS software of Paparazzi enables UAV configuration, monitoring, and custom flight plan configuration for navigational control and guidance. The supported C2 link protocols are MavLink, XBee, SBus, and \ac{PPM}. Paparazzi supports all autonomous navigation features offered by ArduPilot and PX4 in addition to automatic takeoff and landing. 

Several other open-source flight controller software worth mentioning are OpenPilot \cite{openpilot}, LibrePilot \cite{librepilot}, BetaFlight \cite{beta}, dRonin \cite{dronin}, and INAV \cite{inav}. 

\subsection{Computational Unit}

The computational resources on the UAV is a primary concern when it comes to deploying deep learning solutions. The payload capacity of the UAV and the power consumption of the processors are the two major determinants for onboard UAV processor selection. Further, given two processor platforms of comparable weight, an essential performance metric for selection could be the ratio of the inference speed of the deep learning solution to the power consumption of the processor. Additional metrics for selection could be the memory space and volume of the processors.

There are several computational platforms such as  Raspberry Pi 4 Model B, Odroid XU4, Jetson Tegra K1 , SnapDragon flight board, Jetson TX1, among others with on-chip \acp{CPU} and \acp{GPU}. Table \ref{tab:comp} shows a comparison of these platforms in terms of various metrics such as memory, \ac{CPU}, \ac{CPU} speed, \ac{GPU}, \ac{GPU} performance, and dimensions. 

Raspberry Pi 4 Model B (Pi 4B) is a small, low-cost 1.5GHz 64-bit ARM Cortex-A72 \ac{CPU}-based hardware platform with multiple \ac{RAM} options developed for educational purpose. The Pi is also equipped with a Broadcom VideoCore VI \ac{GPU}. However, the Pi 4 model B has a very high power draw in contrast to its predecessors.

Odroid XU4 is a developmental platform that is based on Samsung Exynos 5422 Octa-core \ac{CPU} and ARM Mali-T628 6 Core \ac{GPU}. The XU4 consists of two sockets with 1.4GHz ARM Cortex-A7 and 2GHz ARM Cortex-A15 \acp{CPU}. The Mali-T628 supports OpenGL ES 3.1/2.0/1.1 \cite{opengl} and OpenCL 1.2 \cite{opencl} full profile.

Jetson Tegra K1 (TK1) is a developmental kit from NVIDIA comprising of Kepler \ac{GPU} with 192 CUDA cores and 4-Plus-1 quad-core ARM Cortex-A15 \ac{CPU}. The TK1 has a very low power footprint while being capable of 300 Giga\ac{FLOPS} of 32-bit floating-point computations. The Jetson TX1 on the other hand hosts an NVIDIA Maxwell 256 CUDA core GPU and quad-core ARM Cortex-A57 CPU. The power draw for a typical CUDA load is in the range of 8-10W. In contrast to TK1, the TX1 comes at a much lower form factor of 50mm $\times$ 80mm.

The Snapdragon flight board based on Snapdragon 801 processor was introduced by Qualcomm for autonomous vehicle platforms. The board comes with a 2.26GHz Qualcomm Krait quad-core \ac{CPU} and Qualcomm Adreno 330 \ac{GPU} with nearly 148 Giga\ac{FLOPS} and 4GB \ac{RAM}. In contrast to TX1 and TK1, the Snapdragon flight board comes at a smaller form factor of 58mm $\times$ 40mm. Such a smaller form factor (nearly half the size of a credit card) and lightweight ($<$13g) would serve as an ideal payload option for \acp{UAV}.

\begin{table*}[!h]
\caption{Computational Platforms for \ac{UAV}}
\centering
\def\arraystretch{1.5}%
\begin{tabular}{|p{1.6cm}|p{2.4cm}|p{3cm}|p{2.1cm}|p{2cm}|}
\hline
\textbf{Platforms}       & \textbf{CPU}  & \textbf{GPU}      & \textbf{Dimensions} & \textbf{Memory}\\ \hline
\textbf{Pi 4B} &ARM Cortex-A72 \newline Speed: 1.5GHz &Broadcom VideoCore VI \newline 32Giga\ac{FLOPS} &85mm$\times$56mm &RAM Options: 2GB, 4GB, 8GB\\ \hline
\textbf{Odroid XU4} &ARM Cortex-A7\newline Speed: 1.4GHz \newline ARM Cortex-A15\newline Speed: 2GHz &ARM Mali-T628\newline 102.4Giga\ac{FLOPS}&83mm$\times$59mm &2GB RAM \newline eMMC5.0 HS400 Flash\\ \hline
\textbf{Jetson TK1} &ARM Cortex-A15 \newline Speed: 2.3GHz &Kepler 192 CUDA core\newline 300Giga\ac{FLOPS} &127mm$\times$127mm &2GB RAM \newline 16GB Flash\\ \hline
\textbf{Jetson TX1} &ARM Cortex-A57\newline Speed: 2GHz &Maxwell 256 CUDA core\newline 1Tera\ac{FLOPS} &50mm$\times$87mm &4GB RAM \newline 16GB Flash\\ \hline
\textbf{Snapdragon Flight} &Qualcomm Krait 400 \newline Speed: 2.26GHz &Qualcomm Adreno 330\newline 148Giga\ac{FLOPS}&58mm$\times$40mm &2GB \ac{RAM}\newline 32GB Flash\\ \hline
\end{tabular}
\\
\label{tab:comp}
\end{table*}

Here, we briefly discussed a few computational platforms that could potentially enable deep learning solutions on \ac{UAV} platforms and contrasted them on the basis of their physical and performance specifications. Next, we will discuss the UAS safety and regulations enforced to prevent risk and/or injury to people and property.

\subsection{UAS Safety and Regulations}
\subsubsection*{Safety}
\label{subsec:safety}
\acp{UAV} have become increasingly popular recently for a diverse array of applications including but not limited to personal hobby, photography, aerial survey, precision agriculture, power-line inspection, entertainment, tactical surveillance, border security, etc. \ac{FAA} estimates an even increased adoption of \acp{UAV} in the coming years with an estimate of nearly 3.5 Million in 2021 \cite{faaest}. The advent of \acp{UAV} have posed significant safety and security challenges. Safety encompasses physical risks posed to people and infrastructure as well as UAV cyber-security risks. \ac{FAA} has reported over 4889 incidents causing serious harm to people and infrastructure between 2014 and 2017 \cite{droneincident}. UAV risk factors such as obstacle collision, human factor, rogue \acp{UAV}, untimely battery, and sensor errors, etc., must be carefully assessed prior to any \ac{UAV} missions. Such risk assessment becomes increasingly essential when opting for self-designed \acp{UAV} as opposed to commercial drones. As discussed in section \ref{sec:flightstack}, most of the commercial drones incorporate general safety measures as part of the flight controller software such as obstacle avoidance, return home or land when battery low or sensor error, geofence, among others. Hence, strict \ac{UAV} safety assessment must be conducted in a studied and regulatory manner to alleviate risks to the mission as well as people and infrastructure.

\subsubsection*{Regulations}
In the United States, \ac{FAA} is the regulatory body that enforces aviation rules for air traffic control. Commercial as well as hobbyist use of \acp{UAV} must abide by the regulations enforced by \ac{FAA} as detailed in \cite{faadronezone}. The rules and regulations are enforced based on weight, coverage distance, application, speed, and flight altitude. The regulations restrict operating \acp{UAV} over/near people, in certain airspaces (airports, military facilities, or no-fly zones), and non-line-of-sight operation to avoid accidents and injuries. Commercial \ac{UAV} operation requires the pilots to get licenses as well as are restricted to operate during daylight hours. Recreational flying involves similar rules such as registering the \ac{UAV}, line-of-sight operation, daylight operation, drone altitude not more than 400 feet from the ground, restricted from operating near manned aircraft, people, automobiles, and mental as well as physical alertness during drone operation.
\section{Conclusion}
\label{sec:Conclusion}
This chapter presented how the modern era of machine learning can overcome challenges and accelerate the realization of truly autonomous \ac{UAS}. We begin by presenting a tutorial study of the basic deep learning and reinforcement learning techniques to refine the reader's perception and equip them for further research in this realm. Next, the recent advances in deep learning and reinforcement learning techniques as applied to various autonomous \ac{UAV} tasks were reviewed in depth. The inherent challenges and open problems pertaining to the application of machine learning techniques for autonomous \ac{UAS} tasks were clearly stated to open doors for future research. Additionally, to bridge the gap between simulations and hardware implementations, we present a detailed account of the various simulation suites, \ac{UAV} platforms, flight stacks, and regulatory standards. The various challenges and factors to consider while prototyping \ac{UAV} for machine learning solutions were also discussed. Furthermore, this chapter will serve as a comprehensive handbook to pave a clear roadmap for future research and development in pursuing autonomous \ac{UAS} solutions.

%
%
%
%

%
%

\extrachap{Acronyms}



\begin{acronym}
\acro{AI}{Artificial Intelligence}
\acro{AGL}{Above Ground Level}
\acro{ANN}{Artificial Neural Network}
\acro{CPU}{Central Processing Unit}
\acro{COTS}{commercial-off-the-shelf}
\acro{CAGR}{Compound Annual Growth Rate}
\acro{CNN}{Convolutional Neural Network}
\acro{CRNN}{Convolutional Recurrent Neural Network}
\acro{D3QN}{Dueling Double Deep Q-Network}
\acro{DART}{Dynamic Animation and Robotics Toolkit}
\acro{DPG}{Determinstic Policy Gradient}
\acro{DDPG}{Deep Deterministic Policy Gradient}
\acro{DoD}{Department of Defense}
\acro{DPG}{Deterministic Policy Gradient}
\acro{DQN}{Deep Q-Network}
\acro{DRL}{Deep Reinforcement Learning}
\acro{EKF}{Extended Kalman Filter}
\acro{ESC}{Event Sound Classification}
\acro{FALA}{Finite Action-set Learning Automata}
\acro{FLOPS}{Floating Point Operations Per Second}
\acro{FVI}{Fitted Value Iteration}
\acro{GPU}{Graphics Processing Unit}
\acro{GCS}{Ground Control Station}
\acro{GMM}{Gaussian Mixture Model}
\acro{GPS}{Global Positioning System}
\acro{IMU}{Inertial Measurement Unit}
\acro{IR}{Infrared}
\acro{ISR}{Intelligence, Surveillance and Reconnaissance}
\acro{LSTM}{Long Short Term Memory}
\acro{LWLR}{Locally Weighted Linear Regression}
\acro{MBRL}{Model Based Reinforcement Learning}
\acro{MSL}{Mean Sea Level}
\acro{MAV}{Micro Aerial Vehicle}
\acro{MDP}{Markov Decision Process}
\acro{MPC}{Model Predictive Control}
\acro{NUI}{Natural User Interface}
\acro{ODE}{Open Dynamics Engine}
\acro{OGRE}{Object-Oriented Graphics Rendering Engine}
\acro{PPM}{Pulse Position Modulation}
\acro{PPO}{Proximal Policy Optimization}
\acro{PID}{Proportional-Integral-Derivative}
\acro{PS-DNN}{Partially Shared-Deep Neural Network}
\acro{RAM}{Random Access Memory}
\acro{RDPG}{Recurrent Deterministic Policy Gradient}
\acro{RNN}{Recurrent Neural Network}
\acro{RL}{Reinforcement Learning}
\acro{ROS}{Robot Operating System}
\acro{SARSA}{State-action-reward-state-action}
\acro{SDF}{Simulation Description Format}
\acro{SLAM}{Simultaneous Localization and Mapping}
\acro{STARMAC}{Stanford Testbed of Autonomous Rotorcraft for Multi-Agent Control}
\acro{STFT}{Short Time Fourier Transform }
\acro{SGD}{Stochastic Gradient Descent}
\acro{SWaP}{Size, Weight, and Power}
\acro{TD}{Temporal Difference}
\acro{TRPO}{Trust Region Policy Optimization}
\acro{UAS}{Unmanned Aerial System}
\acro{UAV}{Unmanned Aerial Vehicle}
\acro{UCT}{Upper Confidence bounds applied to Trees}
\acro{VTOL}{Vertical Takeoff and Landing}
\acro{SDK}{Software Development Kit}
\acro{API}{Application Programming Interface}
\acro{RTF}{Ready-to-Fly}
\acro{FAA}{Federal Aviation Administration}
\acro{AMA}{Academy of Model Aeronautics}
\acro{GPL}{GNU General Public License}
\acro{BSD}{Berkeley Software Distribution}
\acro{MAVLink}{Micro Air Vehicle Communication Protocol}
\acro{GUI}{Graphical User Interface}

\end{acronym}

\bibliographystyle{spmpsci}
\bibliography{Reference}
\end{document}